\documentclass[journal]{IEEEtran}

\usepackage{overpic}
\usepackage{graphicx}
\usepackage{amssymb,amsmath,epsfig}
\usepackage{algorithm}
\usepackage{algorithmic}
\usepackage{booktabs}
\usepackage{mathrsfs}
\usepackage{color}
\usepackage{textcomp}
\usepackage{bbding}
\usepackage{pifont}
\usepackage{wasysym}
\usepackage{amssymb}
\usepackage{subcaption}
\usepackage{cite}
\usepackage{amsmath,amssymb,amsfonts}
\usepackage{xcolor}
\usepackage{multirow}
\usepackage{epsfig}
\usepackage[colorlinks, 
            linkcolor=red, 
            anchorcolor=blue, 
            citecolor=green
            ]{hyperref}
\usepackage[switch]{lineno}

\begin{document}
%
\title{Attention-Guided Multiscale Interaction Network for Face Super-Resolution }
%
%
%
\author{Xujie Wan,
        Wenjie Li,
        Guangwei Gao,~\IEEEmembership{Senior Member,~IEEE,}
        Huimin Lu,~\IEEEmembership{Senior Member,~IEEE,}
        Jian Yang,~\IEEEmembership{Member,~IEEE,}
        and Chia-Wen Lin,~\IEEEmembership{Fellow,~IEEE}
\thanks{This work was supported in part by the foundation of the Key Laboratory of Artificial Intelligence of Ministry of Education under Grant AI202404.~\textit{(Xujie Wan and Wenjie Li contributed equally to this work.) (Corresponding author: Guangwei Gao.)}}
\thanks{Xujie Wan and Guangwei Gao are with the Institute of Advanced Technology, Nanjing University of Posts and Telecommunications, Nanjing 210046, China, and also with the Key Laboratory of Artificial Intelligence, Ministry of Education, Shanghai 200240, China (e-mail: wanxujie991205@gmail.com, csggao@gmail.com).}
\thanks{Wenjie Li is with the School of Artificial Intelligence, Beijing University of Posts and Telecommunications, Beijing 100080, China (e-mail: lewj2408@gmail.com).}
\thanks{Huimin Lu is with the School of Automation, Southeast University, Nanjing 210096, China (e-mail: dr.huimin.lu@ieee.org).}
\thanks{Jian Yang is with the School of Computer Science and Engineering, Nanjing University of Science and Technology, Nanjing 210094, China (e-mail: csjyang@njust.edu.cn).}
\thanks{Chia-Wen Lin is with the Department of Electrical Engineering and the Institute of Communications Engineering, National Tsing Hua University, Hsinchu 300044, Taiwan (e-mail: cwlin@ee.nthu.edu.tw).}
}

\markboth{IEEE Transactions on Systems, Man and Cybernetics: Systems}%
{Shell \MakeLowercase{\textit{et al.}}: Bare Demo of IEEEtran.cls for IEEE Journals}
%

\maketitle

\begin{abstract}
Recently, CNN and Transformer hybrid networks demonstrated excellent performance in face super-resolution (FSR) tasks. Since numerous features at different scales in hybrid networks, how to fuse these multiscale features and promote their complementarity is crucial for enhancing FSR. However, existing hybrid network-based FSR methods ignore this, only simply combining the Transformer and CNN. To address this issue, we propose an attention-guided Multiscale interaction network (AMINet), which incorporates local and global feature interactions, as well as encoder-decoder phase feature interactions. Specifically, we propose a Local and Global Feature Interaction Module (LGFI) to promote the fusion of global features and the local features extracted from different receptive fields by our Residual Depth Feature Extraction Module (RDFE). Additionally, we propose a Selective Kernel Attention Fusion Module (SKAF) to adaptively select fusions of different features within the LGFI and encoder-decoder phases. Our above design allows the free flow of multiscale features from within modules and between the encoder and decoder, which can promote the complementarity of different scale features to enhance FSR. Comprehensive experiments confirm that our method consistently performs well with less computational consumption and faster inference.

\end{abstract}


%
\IEEEpeerreviewmaketitle

\section{Introduction}
\label{sec1}

\IEEEPARstart{F}{ace} super-resolution (FSR), also known as face hallucination, aims at restoring high-resolution (HR) face images from low-resolution (LR) face images~\cite{liu2020discriminative}. In contrast to standard image super-resolution, the primary objective of FSR is to reconstruct as many facial structural features as possible (\emph{i.e.} the shape and contour of facial components). In practical scenarios, a range of face-specific tasks, such as face detection~\cite{mamieva2023improved} and face recognition~\cite{hu2015face}, require HR face images. However, the quality of captured face images is frequently diminished due to variations in hardware configuration, positioning, and shooting angles of the imaging devices, seriously affecting the above downstream tasks. Therefore, FSR has garnered increasing attention in recent years.

\begin{figure}[t]
\centering
\includegraphics[width=9cm, trim=0 0 30 40]{img/FIG.png}
\caption{Model complexity studies for $\times$ 8 FSR on CelebA test set~\cite{liu2015deep}. Our AMINet achieves an excellent balance between model size, model performance, and inference speed. }
\label{fig: Model_Complexity}
\end{figure}

Recently, since the advantages exhibited by hybrid networks~\cite{gao2023ctcnet} of CNNs and Transformers in FSR, this type of method has gained increased attention. Specifically, CNN-based FSR methods~\cite{zhou2015learning} generally do not require large computational costs. Still, they specialize in extracting local details, such as the local texture of the face, color, \emph{etc.}, and are unable to model long-range feature interaction, such as the global profile of the face. Transformer-based FSR methods~\cite{shi2022idpt} can simulate global modeling well, but their computational consumption is huge. Hybrid FSR methods leverage the strengths of both architectures, enabling efficient extraction of both local and global features while maintaining a reasonable computational overhead. The impressive performance of hybrid-based FSR methods comes from numerous features extracted inside their networks at different scales, such as global features from self-attention, local features from convolution, and features from different stages of the encoder-decoder, which facilitates models to refine local details and global contours.

However, while existing hybrid-based FSR methods consider utilizing features from different scales to improve FSR, they ignore the problem of how to fuse these multiscale features to make their properties better complement each other. For example, Faceformer~\cite{wang2022faceformer} simply parallelizes the connected CNN modules and the window-based Transformer~\cite{liang2021swinir} modules. SCTANet~\cite{bao2023sctanet} also only juxtaposes spatial attention-based residual blocks and multi-head self-attention in designed modules. CTCNet~\cite{gao2023ctcnet} simply connects the CNN module in tandem with the Transformer module. These methods overlook the importance of blending multiscale features in a complementary manner and enabling smooth information flow across different scales to refine facial details effectively.

To address this problem, we propose an Attention-Guided Multiscale Interaction Network (AMINet) for FSR in this work. Our AMINet fuses multiscale features in two main ways, including the fusion of features obtained from self-attention and convolution, and the fusion of features at different stages of the encoder-decoder. Specifically, we design a Local and Global Feature Interaction Module (LGFI) to adaptively fuse global facial and local features with different receptive fields obtained by convolutions. In LGFI, self-attention is responsible for extracting global features, while our Residual Depth Feature Extraction Module (RDFE) extracts local features at different scales using separable convolutional kernels of different sizes, and our Selective Kernel Attention Fusion Module (SKAF) is responsible for weighted fusion of these two parts of features for our model to adaptively perform selective fusion during training. In addition, we also utilize our SKAF as a crucial fusion module in our Encoder and Decoder Feature Fusion Module (EDFF) to further perform feature communications of our method by fusing features at different scales from the encoder-decoder processes. 

Our above design greatly enhances the flow and exchange of features at different scales within the model and improves the representation of our model. As a result, our method can obtain a more powerful feature representation than existing FSR methods. As shown in Fig.~\ref{fig: Model_Complexity}, our method can achieve the best FSR performance with a smaller size and faster inference speed, demonstrating our method's effectiveness. In summary, the main contributions are as follows:

\begin{itemize}
\item We design an LGFI to differ from the traditional Transformer by allowing free flow and selective fusion of local and global features within the module.

\item We design an RDFE, which enables better refinement of facial details by fusion and refinement of local features extracted by convolutional kernels of different sizes.

\item We design the SKAF to help selective fusions of different-scale features within LGFI and EDFF by selecting appropriate convolutional kernels.

\end{itemize}

\begin{figure*}[t]
	\centerline{\includegraphics[width=17.8cm]{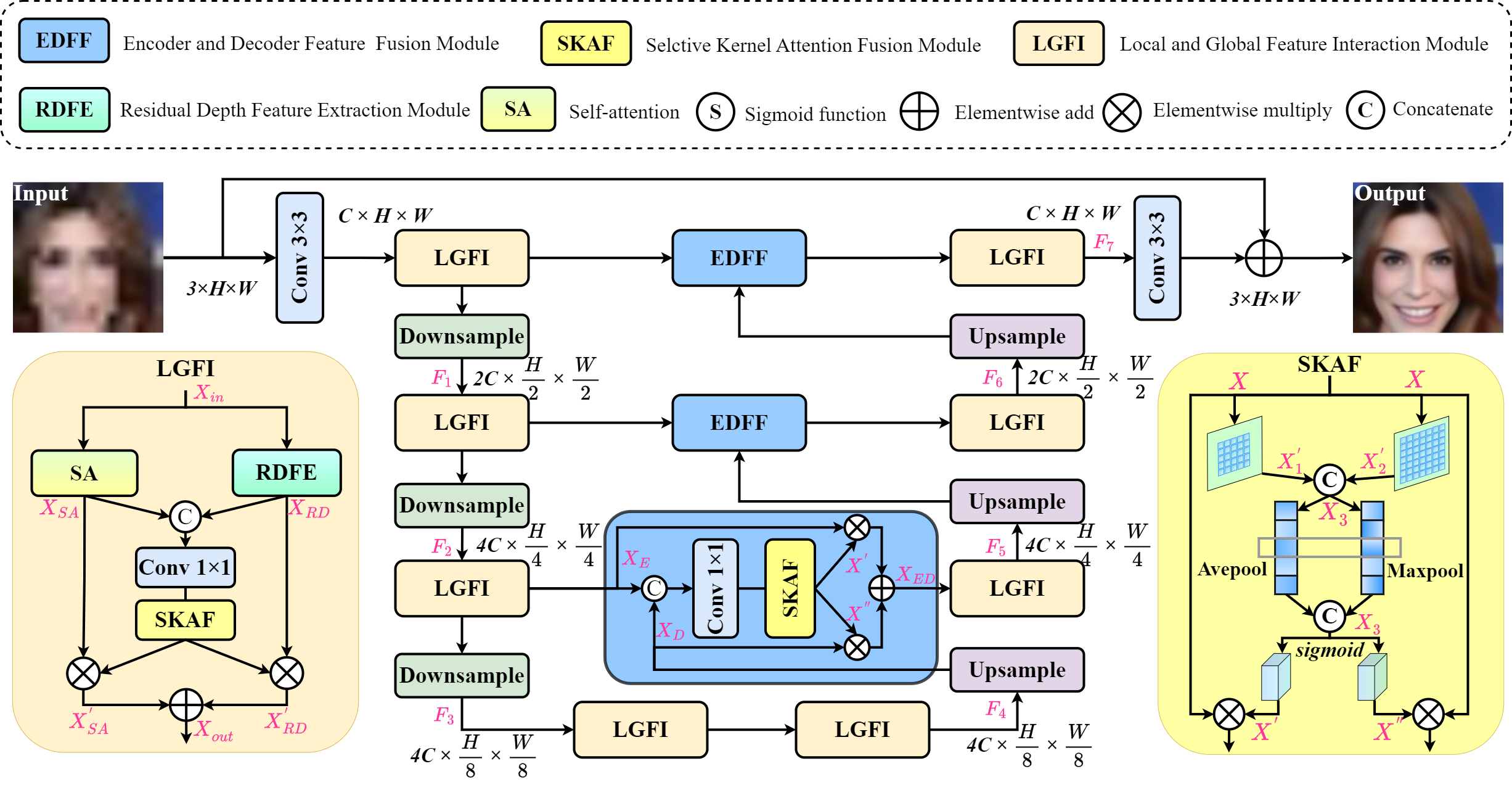}}
	\caption{Network structure of our AMINet, which is a U-shaped CNN-Transformer hierarchical architecture with three distinct stages: encoding, bottleneck, and decoding.}
	\label{Network structure}
\end{figure*}




\section{Related Work}
\label{sec2}

\subsection{Face Super-Resolution}
\label{sec21}
Early deep learning approaches focused on leveraging facial priors as guidance to enhance FSR accuracy~\cite{li2023survey}. For instance, Chen \textit{et al.}~\cite{chen2018fsrnet} developed an end-to-end prior-based network that utilized facial landmarks and heatmaps to generate FSR images. Similarly, Kim \textit{et al.}~\cite{kim2019progressive} employed a face alignment network for landmark extraction in conjunction with a progressive training technique to produce realistic face images. Ma \textit{et al.}~\cite{ma2020deep} introduced DICNet, which iteratively integrates facial landmark priors to enhance image quality at each step. Hu \textit{et al.}~\cite{hu2020face} explored the use of 3D shape priors to better capture and define sharp facial structures. While these methods have advanced FSR, they require additional labeling of training datasets. Moreover, inaccuracies in prior estimation can significantly diminish FSR performance, especially when dealing with highly blurred face images.

Attention-based FSR methods have been proposed to promote FSR to avoid the adverse effects of inaccurate prior estimates on FSR. Zhang~\textit{et al.}~\cite{zhang2020supervised} proposed a supervised pixel-by-pixel generation of the adversarial network to improve face recognition performance during FSR. Chen~\textit{et al.}~\cite{chen2020learning} proposed the SPARNet, which can focus on important facial structure features adaptively by using spatial attention in residual blocks. Lu~\textit{et al.}~\cite{lu2021face} proposed a partial attention mechanism to enhance the consistency of the fidelity of facial detail and facial structure. Bao~\textit{et al.}~\cite{bao2022distilling} introduced the equalization texture enhancement module to enhance the facial texture detail through histogram equalization. Wang \textit{et al.}~\cite{wang2023spatial} critically introduced the Fourier transform into FSR, fully exploring the correlation between spatial domain features and frequency domain features. Shi~\textit{et al.}~\cite{shi2023exploiting} designed a two-branch network, which introduces a convolution based on local changes to enhance the ability of the convolution. Liu~\textit{et al.}~\cite{bao2023sctanet} improves the interaction ability of regional and global features through designed hybrid attention modules. Li \textit{et al.}~\cite{li2024efficient} designed a wavelet-based network to reduce the loss of downsampling in the encoder-decoder. Although the above methods can reconstruct reasonable FSR images, they cannot promote the efficient fusion of local features with global features and different features at different stages of the encoder-decoder, affecting FSR's efficiency and accuracy.

\subsection{Attention-based Super-Resolution}
The attention mechanism can improve the super-resolution accuracy of models due to its flexibility in focusing on key areas of facial features. In the super-resolution task, different variants of the attention mechanism include self-attention, spatial attention, channel attention, and hybrid attention~\cite{li2024systematic}.

Zhang \textit{et al.}~\cite{zhang2018image} inserted channel attention into residual blocks to enhance model representation. Xin \textit{et al.}~\cite{xin2019residual} utilized channel attention plus residual mechanisms to combine a multi-level information fusion strategy. Chen \textit{et al.}~\cite{chen2020learning} enhanced FSR by utilizing an improved facial spatial attention that cooperated with the hourglass structure. Gao \textit{et al.}~\cite{gao2022feature} performed shuffling to hybrid attention.  Wang \textit{et al.}~\cite{wang2024multi} constructed a simplified feed-forward network using spatial attention to reduce parameters and computational complexity. To model long-range feature interaction, the self-attention in Transformer~\cite{devlin2018bert} has been widely used in super-resolution. Gao \textit{et al.}~\cite{gao2022lightweight} reduced costs by utilizing the recursive mechanism on self-attention. Li~\textit{et al.}~\cite{li2023cross,li2022efficient} combined self-attention and convolutions to complement each other's required features. Zeng \textit{et al.}~\cite{zeng2023self} introduced a self-attention network that investigates the relationships among features at various levels. Shi \textit{et al.}~\cite{shi2023exploiting} enhanced FSR by mitigating the adverse effects of inaccurate prior estimates through a parallel self-attention mechanism, effectively capturing both local and non-local dependencies. To combine the advantages of different attentions, Yang \textit{et al.}~\cite{yang2021image} integrated channel attention with spatial attention to enhance feature acquisition and correlation modeling. Bao \textit{et al.}~\cite{bao2023sctanet} and Gao \textit{et al.}~\cite{gao2023ctcnet} employed spatial attention and self-attention to capture facial structure and details. Zhang \textit{et al.}~\cite{zhang2024feature} employed a hybrid attention that combines self-attention, spatial attention, and channel attention to optimize fine-grained facial details and broad facial structure. Unlike the above attention-based methods that enhance model representation, we utilize attention to learn features from different receptive fields, allowing our network to adaptively select the appropriate convolutional kernel size to match the multiscale feature fusion. This design enables our network to perform multiscale feature extraction and improve the integration of features across various scales, leading to enhanced performance and greater adaptability.





\section{Proposed Method}
\label{sec3}


\begin{figure*}[t]
	\centerline{\includegraphics[width=17.8cm]{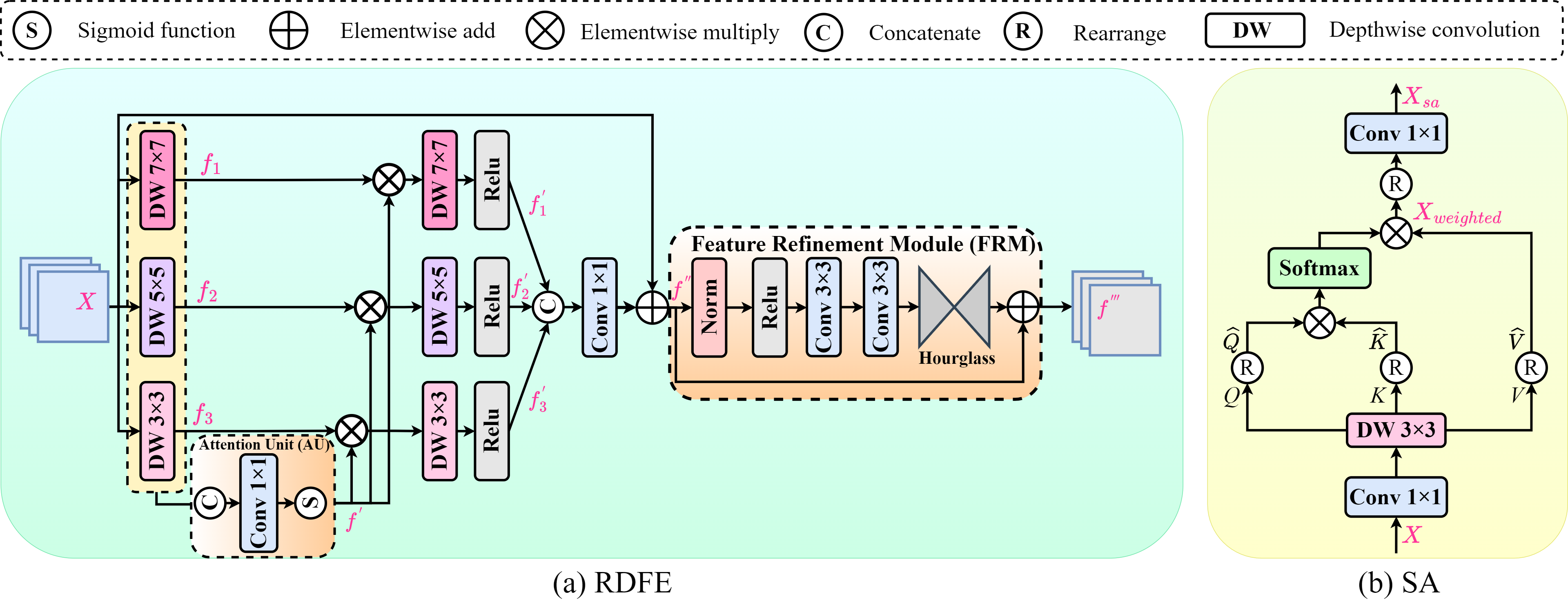}}
	\caption{ The architecture of (a) Residual Depth Feature Extraction Module (RDFE), (b) Self-attention (SA), respectively.}
	\label{RDFEM and SA structure}
\end{figure*}

\subsection{Overview of AMINet}
\label{sec31}

As illustrated in Fig.~\ref{Network structure}, our proposed AMINet features a U-shaped CNN-Transformer hierarchical architecture with three distinct stages: encoding, bottleneck, and decoding. For an LR input face image ${I_{LR} \in \mathbb{R}^{3 \times H \times W}}$, in the encoding stage, our network aims to extract features at different scales and capture multiscale feature representations of the input image to get the facial feature ${F_{3} \in \mathbb{R}^{C \times H \times W}}$. Then, the bottleneck stage network continues to refine the feature ${F_{3}}$ and provides a more informative representation to get the refined feature ${F_{4}}$ for the subsequent reconstruction phase. In decoding, the network focuses on feature upsample and facial detail reconstruction. Meanwhile, an interactive connection is used between the encoding and decoding stages to ensure the features are fully integrated throughout the network. We can get the reconstructed face feature ${F_{7}}$ with rich facial details through the above operators. Finally, through a convolution with reduced channel dimensions plus a residual connection, we get the HR output face image ${I_{HR} \in \mathbb{R}^{3 \times H \times W}}$.


\subsubsection{Encoding stage} The encoding stage in our network aims to extract facial features of different scales. In this stage, given a degraded face image ${I_{LR} \in \mathbb{R}^{3 \times H \times W}}$, first, a ${3\times3}$ convolution is used to extract shallow facial features. Then, extracted facial features are further refined by three encoder stages. Each encoder comprises our designed Local and Global Feature Interaction Module (LGFI) and a downsampling operator. After each encoder, the input face feature's channel count will be doubled, and the size of the image of the input face feature will be halved. As shown in Fig.~\ref{Network structure}, the features obtained after three encoders are as follows: ${F_{1} \in \mathbb{R}^{2C \times \frac{H}{2} \times \frac{W}{2}}}$, ${F_{2} \in \mathbb{R}^{4C \times \frac{H}{4} \times \frac{W}{4}}}$, ${F_{3} \in \mathbb{R}^{8C \times \frac{H}{8} \times \frac{W}{8}}}$.

\subsubsection{Bottleneck stage} In the bottleneck stage between the encoding and decoding stages, the obtained encoding features are designed to be fine-grained. ${F_{4} \in \mathbb{R}^{8C \times \frac{H}{8} \times \frac{W}{8}}}$ is obtained through the bottleneck stage. In this stage, we continue to use two LGFIs to refine and enhance encoding features to ensure they are better utilized in the decoding stage. After this stage, our model can continuously enhance the information about the facial structure at different scales, thus improving the perception of facial details.

\subsubsection{Decoding stage} In the decoding stage, there are three decoders. We focus on multiscale feature fusion, aiming at reconstructing high-quality face images at this stage. As depicted in Fig.~\ref{Network structure}, each decoder includes an upsampling operation, an EDFF, and an LGFI. Each upsampling operator halves the input feature channel counts while doubling the width and weight of the input facial feature. Compared to encoding stages, decoding stages additionally use our proposed SKAF to adaptively and selectively fuse different scale features from the encoder and decoder stages. Through this design, different scale features can interact to recover more detailed face features. The features obtained after three decoders are as follows: ${F_{5} \in \mathbb{R}^{4C \times \frac{H}{4} \times \frac{W}{4}}}$, ${F_{7} \in \mathbb{R}^{C \times H \times W}}$. Finally, a ${3\times3}$ convolution unit is utilized to transform our obtained deep facial feature into an output FSR image ${I_{SR}}$.

\subsection{ Local and Global Feature Interaction Module (LGFI)}
\label{sec32}

In our AMINet, LGFI is mainly used for local and global facial feature extraction. As shown in Fig.~\ref{Network structure}, LGFI consists of  Self-attention(SA), a Residual Depth Feature Extraction Module (RDFE), and a Selective Kernel Attention Fusion Module (SKAF), used for local and global feature fusion and interaction, respectively. The SA is designed to extract global features. At the same time, RDEM is designed to extract local features at different scales and enrich local facial details through multiple convolutional kernels under numerous receptive fields. Specifically, the integration of local and global features is achieved through SKAF, which adaptively weights and fuses multiscale information. SKAF first extracts local and global features via convolutions with different receptive fields and then computes their importance using average pooling. These computed weights are applied to the corresponding features, facilitating an adaptive fusion process. By dynamically selecting and emphasizing key information, SKAF enhances the complementarity between local textures and global structural cues, ultimately improving facial detail reconstruction and overall super-resolution performance.

\subsubsection{Self-attention (SA)} We utilize Self-attention (SA) to extract global facial features, which can effectively model the relationships between distant features. Meanwhile, through the multi-head mechanism in SR, features can be captured from different subspaces, improving the robustness and generalization ability of the model. As illustrated in Fig.~\ref{RDFEM and SA structure} (b), we start by applying a ${1\times1}$ convolutional layer followed by a ${3\times3}$ depth-wise convolutional layer to combine pixel-level cross-channel information and extract channel-level spatial context. From this spatial context, we then generate ${Q, K, V \in \mathbb{R}^{C\times H\times W}}$. For an input facial feature $X\in \mathbb{R}^{C\times H\times W}$ , the process of obtaining ${Q,K,V\in \mathbb{R} ^{C\times H\times W}}$ can be described as:
\begin{equation}
{Q = F_{dw3}(F_{conv1}(X))},
\end{equation}
\begin{equation}
{K = F_{dw3}(F_{conv1}(X))},
\end{equation}
\begin{equation}
{V = F_{dw3}(F_{conv1}(X))},
\end{equation}
where ${F_{conv1}}(\cdot)$ is the ${1\times1}$ pointwise convlution and ${F_{dw3}}(\cdot)$ is the ${3\times3}$ depthwise convlution.

Next, we reshape $Q$, $K$, and $V$ into ${\hat{Q} \in \mathbb{R}^{C \times HW}}$, ${\hat{K} \in \mathbb{R}^{HW \times C}}$, and ${\hat{V} \in \mathbb{R}^{C \times HW}}$, respectively. After that, the dot product is multiplied by $V$ to obtain weights $X_{w}\in \mathbb{R}^{C\times HW}$, which facilitates the capturing of the important local context in SA. Finally, we rearrange $X_{w}$ into $\hat{X_{w}}\in \mathbb{R}^{C\times H\times W}$. The above operations can be expressed as:
\begin{equation}
{X_{weighted} = \operatorname{Softmax}(\hat{Q}\cdot\hat{K}/\sqrt{d})} \cdot \hat{V},
\end{equation}
\begin{equation}
{X_{sa} = F_{conv1}(R(X_{weighted}))},
\end{equation}
where ${X_{sa}}$ is the attention map of SA, ${\sqrt{d}}$ is a factor used to scale the dot product of ${\hat{K}}$ and ${\hat{Q}}$, ${R(\cdot)}$ stands for the rearrange operation, ${X_{sa}}$ denotes the output of SA.

\subsubsection{Residual depth feature extraction module (RDFE)} As shown in Fig.~\ref{RDFEM and SA structure} (a), we design RDFE to extract local facial features at different scales. Compared with the traditional feed-forward network (FFN), our RDFE is beneficial for processing more complex features and multiscale features flexibly. Specifically, for the input feature $\mathop X \in \mathbb{R}^{C\times H\times W}$, we use depthwise convolutions of ${3 \times3} $, ${5 \times5} $, and ${7 \times7} $ to parallelly extract three scales facial features, which depthwise convolution can reduce the computational complexity of the model, while convolution with different kernel sizes can effectively extracting rich face details. The reason we employ depthwise separable convolution is to significantly reduce computational complexity while preserving local spatial features. In our model, depthwise convolutions at different scales enable the extraction of fine-grained facial details, such as textures and edge information, across varying receptive fields, enhancing the model’s ability to capture local features efficiently. The above operations can be expressed as:
\begin{equation}
\mathop f\nolimits_{1},\mathop f\nolimits_{2},\mathop f\nolimits_{3} = \mathop F\nolimits_{dw3}   (\mathop X) ,\mathop F\nolimits_{dw5}   (\mathop X),\mathop F\nolimits_{dw7}   (\mathop X),
\end{equation}
where $F_{dw3}$, $F_{dw5}$, and $F_{dw7}$ are ${3 \times3} $, ${5 \times5} $, and ${7 \times7}$ depthwise convlution, respectively. However, relying solely on convolutions may lead to insufficient feature fusion across scales, potentially introducing redundant information or affecting the representation of key facial details. To address this issue, we use an attention unit $F_{au}$ to calculate the feature weight of the fusion feature of three branches. Next, we use the calculated weights to modulate the three-branch features at different scales through element-wise multiplications. This mechanism enables the model to dynamically balance the contributions of multiscale features, effectively capturing key facial details across varying receptive fields, thereby enhancing detail restoration and overall generalization. To further refine and reconstruct the weighted fused facial features, we apply depthwise convolutions with kernel sizes of ${3 \times3}$, ${5 \times5}$, and ${7 \times7}$, ensuring the joint optimization of local texture details and global structural consistency. This collaborative refinement ultimately leads to higher-quality facial reconstructions. The above operations can be expressed as:
\begin{equation}
\mathop f\nolimits^{'} = \mathop F\nolimits_{au} (\mathop H\nolimits_{cat}(\mathop f\nolimits_{1},\mathop f\nolimits_{2},\mathop f\nolimits_{3})),
\end{equation}
\begin{equation}
\mathop f\nolimits_{1}^{'},\mathop f\nolimits_{2}^{'},\mathop f\nolimits_{3}^{'} = \mathop f\nolimits_{1} (\mathop X)\cdot\mathop f\nolimits^{'} ,\mathop f\nolimits_{2}   (\mathop X)\cdot\mathop f\nolimits^{'},\mathop f\nolimits_{3}   (\mathop X)\cdot\mathop f\nolimits^{'},
\end{equation}
where $H_{cat}$ is a concat operator and $\mathop F\nolimits_{au} ( \cdot )$  indicates attention unit. Then, we aggregate the three branches' features to combine facial detail information under different receptive fields. This process can be described as:
\begin{equation}
\mathop f\nolimits^{''}= \mathop H\nolimits_{cat}(\mathop F\nolimits_{conv1}(\mathop f\nolimits_{1}^{'},\mathop f\nolimits_{2}^{'},\mathop f\nolimits_{3}^{'}))+X
\end{equation}
where $\mathop F\nolimits_{conv1} ( \cdot )$ represents
${1\times1}$ convolution. Finally, we utilize a Feature Refinement Module to refine features obtained from previous multiple branches. Specifically, we begin by applying normalization and multiple $3 \times 3$ convolutional layers to refine the local facial context. Afterward, the hourglass block further integrates multiscale information to capture global and local relationships:
\begin{equation}
\mathop f\nolimits^{'''}= \mathop F\nolimits_{frm}(\mathop f\nolimits^{''}),
\end{equation}
where $\mathop F\nolimits_{frm} ( \cdot )$ indicates feature refinement module.

\subsubsection{Selective Kernel Attention Fusion Module (SKAF)}  
Inspired by SKNet~\cite{li2019selective} and LSKNet~\cite{li2023large}, as shown in Fig.~\ref{Network structure}, we design an SKAF module to give our model the ability to select local and global features required for reconstruction for fusion interaction. Specifically, SKAF first extracts both local and global features using convolutions with different receptive fields. It then computes feature importance through an average pooling operation, which was previously not explicitly shown in the diagram. Finally, an adaptive weighting mechanism selects and fuses the most relevant local and global information, enabling SKAF to dynamically emphasize critical features under varying receptive fields. This design enhances the model’s capacity for feature selection, improving both its performance and generalization in face super-resolution. Given feature $\mathop X\nolimits \in \mathbb{R}^{C\times H\times W}$ obtained by the SA and the RDFE, we first fuse the local and global features extracted by a $5 \times 5$ convolution and a $7 \times 7$ convolution to get a hybrid feature $X$. This operation can be expressed as:
\begin{equation}
\mathop X\nolimits_{1}^{'},\mathop X\nolimits_{2}^{'} = \mathop F\nolimits_{conv5}   (\mathop X\nolimits ),\mathop F\nolimits_{conv7}  (\mathop X\nolimits ),
\end{equation}
\begin{equation}
\mathop X\nolimits_{3}^{} =  \mathop H\nolimits_{cat} (\mathop X\nolimits_{1}^{'} ,\mathop X\nolimits_{2}^{'} ),
\end{equation}
where $\mathop F\nolimits_{conv5} ( \cdot )$  represents
${5\times5}$  convolution, $\mathop F\nolimits_{conv7} ( \cdot )$  represents ${7\times7}$  convolution, $\mathop H\nolimits_{cat} ( \cdot )$ indicates
the concat operation along the channel dimension. Then, we impose pooling to learn the weight of the obtained hybrid features, where the weight reflects the importance of features under different receptive fields. The process of obtaining the weight for selecting required facial features is as follows:
\begin{equation}
\mathop X\nolimits_{3}^{} = \mathop H\nolimits_{sig} (\mathop H\nolimits_{cat} (\mathop H\nolimits_{avep} (\mathop X\nolimits_{3}^{}),\mathop H\nolimits_{maxp} (\mathop X\nolimits_{3}^{}))),
\end{equation}
where $\mathop H\nolimits_{avep} ( \cdot )$, $\mathop H\nolimits_{maxp} ( \cdot )$, and $\mathop H\nolimits_{sig} ( \cdot )$ indicate the average and max pooling operation along the channel direction and sigmoid function, respectively.  Finally, we multiply the weights obtained from the above calculations with the local and global features, respectively. Thus, our SKAF can adaptively select the important local and global information required for reconstruction. The process of obtaining local and global features $X^{'}$, $X^{''}$ by adaptive weight selection is:
\begin{equation}
\mathop X\nolimits^{'},\mathop X\nolimits^{''}  = \mathop H\nolimits_{cs} (\mathop X\nolimits_{3}^{}),
\end{equation}
where $\mathop H\nolimits_{cs} ( \cdot )$ indicates the feature separation operation along the channel dimension. Through the above operators, we can get the adaptive selected local and global features.

\subsection{Encoder and Decoder Feature Fusion Module (EDFF)}
\label{sec33}
To fully utilize the multiscale features extracted from the encoding and decoding stage, we introduce an EDFF to fuse different features, enabling our AMINet with better feature propagation and representation capabilities. As shown in Fig.~\ref{Network structure}, our EDFF mainly utilizes our proposed SKAF to fuse and select different-scale features required for reconstruction. Given the feature $\mathop X\nolimits_E \in \mathbb{R}^{C\times H\times W}$, the feature $\mathop X\nolimits_D \in \mathbb{R}^{C\times H\times W}$ is from the decoding stage and the encoding stage, respectively. Firstly, we concatenate features from the encoding and decoding stages along the channel dimension. Then, a ${1 \times1}$ convolution is used to reduce the channel counts and reduce the process's computational costs to obtain two weights through our SKAF, which can be expressed as:
\begin{equation}
\mathop X\nolimits^{'},\mathop X\nolimits^{''} = \mathop F\nolimits_{skaf} (\mathop F\nolimits_{conv1} (\mathop H\nolimits_{cat} (\mathop X\nolimits_E ,\mathop X\nolimits_D ))),
\end{equation}
where $\mathop F\nolimits_{skaf} ( \cdot )$  represents
Selective Kernel Attention Fusion Module,  $\mathop F\nolimits_{conv1} ( \cdot )$ stands for the ${1\times1}$ convolutional layer, and $\mathop H\nolimits_{cat} ( \cdot )$ denotes the operation of concatenating features across the channel dimension. Next, we feed the two obtained weights into two branches for multiplication. Through this operator, we obtain the selected facial features from hybrid features obtained by fusing the encoding and decoding features. Finally, we add the features of the two branches:
\begin{equation}
\mathop X\nolimits_{ED}  = \mathop X\nolimits_{E}\cdot\mathop X\nolimits^{'}  + \mathop X\nolimits_{D}\cdot\mathop X\nolimits^{''}.
\end{equation}
Through the above operators, we can complete the process of the adaptive fusion of encoding and decoding features.

\textbf{\begin{figure}[t]
\centering
\includegraphics[width=8.5cm, trim=0 0 0 0]{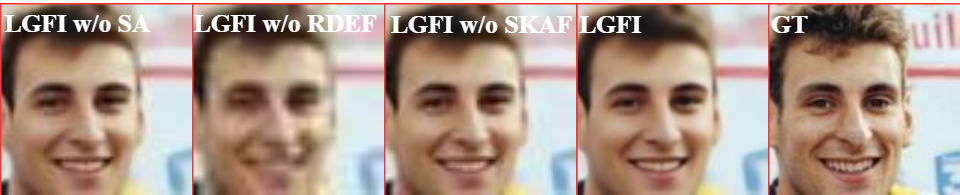}
\caption{Visual comparison of various outputs of LGFI.}
\label{compare_lgfi}
\end{figure}}

\textbf{\begin{figure}[t]
\centering
\includegraphics[width=8.5cm, trim=0 0 0 0]{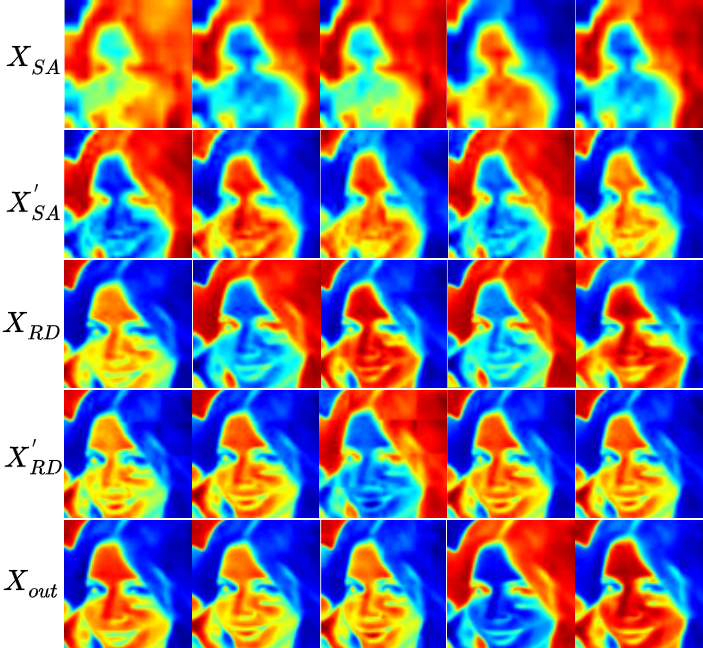}
\caption{Heat map outputs of different parts of LGFI.}
\label{heatmap_lgfi}
\end{figure}}

\begin{table}[t]
	\begin{center}
		\caption{Verify the effectiveness of LGFI \textcolor{black}{(CelebA, $\times 8$)}.}
		\setlength{\tabcolsep}{1.57mm}
        \renewcommand\arraystretch{1}
		\begin{tabular}{c|ccccc}
			\hline
			Methods    & PSNR/SSIM${\uparrow}$   & VIF${\uparrow}$  & LPIPS${\downarrow}$ &Speed${\downarrow}$  & Params${\downarrow}$   \\
			\cline{2-5}
			\hline
			\hline
			LGFI w/o SA      & 27.75/0.7944       &0.4652        &0.1886    &33ms     &12.3M \\
			LGFI w/o RDFE    & 27.51/0.7840       & 0.4495       &0.2085    &21ms     & 7.2M \\
            LGFI w/o SKAF    & 27.74/0.7932       & 0.4611       &0.1979      &25ms   &8.21M  \\
			LGFI             &\bf{27.83/0.7961}  &\bf{0.4725}   &\bf{0.1821}   &34ms  &12.62M\\
			\hline
		\end{tabular}
		\label{Effects of LGFIM}
	\end{center}
\end{table}

\begin{table}[!t]
	\begin{center}
		\caption{Quantatitive comparison between LGFI and traditional Transformer as shown in Fig.~\ref{fig:compare_structure} \textcolor{black}{(CelebA, $\times 8$)}.}
        \setlength{\tabcolsep}{2.5mm}
        \renewcommand\arraystretch{1}
		\begin{tabular}{c|ccccc}
			\hline
    Methods &Parameters &PSNR${\uparrow}$ &SSIM${\uparrow}$ &VIF${\uparrow}$  &LPIPS${\downarrow}$ \\
			\cline{2-5}
			\hline
			\hline
			 Transformer    &11.32M   &27.73    &0.7952   &0.4511   &0.1878  \\
                LGFI     & 12.62M     &\bf{27.83}  &\bf{0.7961}   &\bf{0.4725} &\bf{0.1821}    \\
			\hline
		\end{tabular}
		\label{Comparison between LGfim and Trans}
	\end{center}
\end{table}

\textbf{\begin{figure}[t]
\centering
\includegraphics[width=8cm, trim=0 0 0 0]{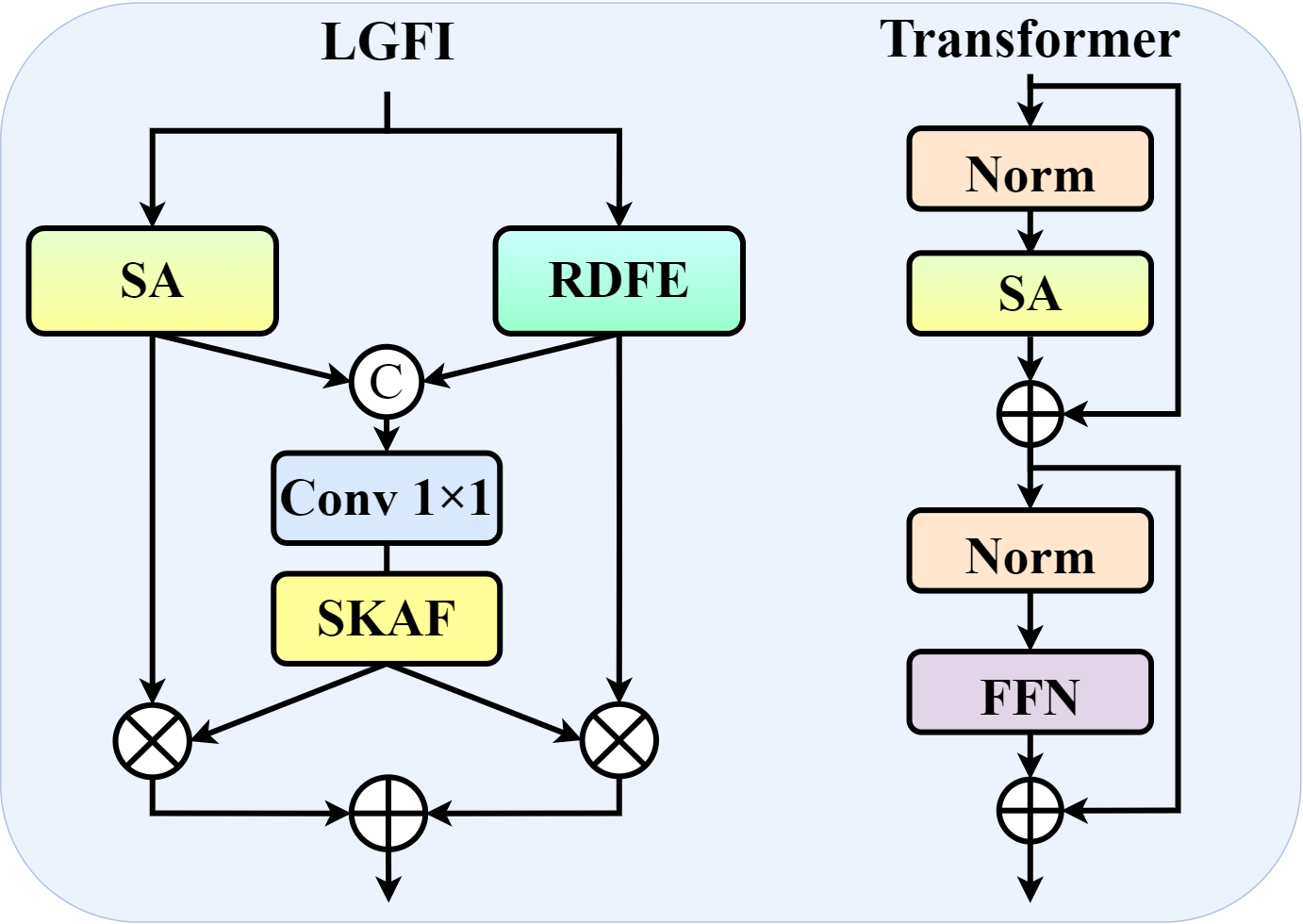}
\caption{Comparison of LGFI and Transformer structures, where SA is self-attention, RDFE, and FFN are CNN parts, and SKAF is our feature fusion module.}
\label{fig:compare_structure}
\end{figure}}

\subsection{Loss Functions}
\label{sec36}

As for the loss of our AMINet, given a dataset $\mathop {\left\{ {\mathop I\nolimits_{LR}^i,\mathop I\nolimits_{HR}^i } \right\}}\nolimits_{i = 1}^N $, we optimize our AMINet by minimizing the pixel-level loss function:
\begin{equation}
{\mathcal{L}(\Theta) = \frac{1}{N}\sum_{i=1}^{N}\left \|F_{AMINet}(I_{LR}^{i},\Theta ) -I_{HR}^{i}\right\|_{1}},
\end{equation}
where $N$ denotes paired training face image counts. ${I_{LR}^{i}}$ and ${I_{HR}^{i}}$ are the face LR image and HR image of the $i$-th pair, respectively. Meanwhile, $F_{AMINet}(\cdot)$ and $\Theta$ denote the AMINet and the number of parameters of AMINet, respectively.

Since the GAN-based methods~\cite{ledig2017photo,wang2018esrgan} can get better perceptual qualities, we expand our AMINet to AMIGAN to generate more high-quality SR results. The loss function used in training AMIGAN consists of the following three parts:

\subsubsection{Pixel loss} Pixel-level loss is used to reduce the pixel difference between the SR and HR images:
\begin{equation}
{\mathcal{L}_{pix}=\frac{1}{N} \sum_{i=1}^{N}\left\|G(I_{LR}^{i})-I_{HR}^{i}\right\|_{1}},
\end{equation}
where $G$ indicates the AMIGAN generator.

\subsubsection{Perceptual loss} To enhance the visual quality of super-resolution images, we apply perceptual loss. This involves using a pre-trained VGG19~\cite{simonyan2014very} model to extract facial features from both the HR images and our generated FSR images. Then, we compare the obtained perceptual features of HR and FSR images to constrain the generation of FSR features. Therefore, the perceptual loss can be described as:
\begin{equation}
{\mathcal{L}_{pcp}=\frac{1}{N}\sum_{i=1}^{N}\sum_{l=1}^{L_{VGG}}\frac{1}{M_{VGG}^{l}}\left\|f_{VGG}^{l}\left(I_{SR}^{i}\right)-f_{VGG}^{l}\left(I_{HR}^{i}\right)\right\|_{1}}, 
\end{equation}
where $f_{VGG}^{l}$ represents the feature map from the $l$-th layer of the VGG network, $L_{VGG}$ is the total number of layers in VGG, and $M_{VGG}^{l}$ indicates the quantity of elements within that feature map.

\subsubsection{Adversarial loss} GANs have been shown to be effective in reconstructing photorealistic images~\cite{ledig2017photo,wang2018esrgan}. GAN generates FSR results through the generator while using the discriminator to distinguish between ground truth and FSR results, which ultimately enables the generator to generate realistic FSR results in the process of constant confrontation. This process is denoted as:
\begin{equation}
{\mathcal{L}_{dis}=-\mathbb{E}\left[\log\left(D\left(I_{H R}\right)\right)\right]-\mathbb{E}\left[\log\left(1-D\left(G\left(I_{L R}\right)\right)\right)\right]}.
\end{equation}
Additionally, the generator tries to minimize: 
\begin{equation}
\mathcal{L}_{adv}=-\mathbb{E}\left[\log \left(D\left(G\left(I_{L R}\right)\right)\right)\right].
\end{equation}
Thus, AMIGAN is refined by minimizing the following total objective function:
\begin{equation}
{\mathcal{L}=\lambda _{pix}\mathcal{L}_{pix}+\lambda_{pcp}\mathcal{L}_{pcp}+\lambda_{adv}\mathcal{L}_{adv}},
\end{equation}
where $\lambda_{pix}$, $\lambda_{pcp}$, and $\lambda_{adv}$ represent the weighting factors for the corresponding pixel loss, perceptual loss, and adversarial loss, respectively.

\begin{table}[t]
	\begin{center}
		\caption{Performance and computational cost comparison between RDFE and FFN \textcolor{black}{(CelebA, $\times 8$)}.}
        \setlength{\tabcolsep}{2.5mm}
        \renewcommand\arraystretch{1}
		\begin{tabular}{c|ccccc}
			\hline
    Methods &Parameters &PSNR${\uparrow}$ &SSIM${\uparrow}$ &VIF${\uparrow}$  &LPIPS${\downarrow}$ \\
			\cline{2-5}
			\hline
			\hline
			 FFN    &12.11M   &27.72    &0.7931   &0.4578   &0.1922  \\
                RDFE     & 12.62M     &\bf{27.83}  &\bf{0.7961}   &\bf{0.4725} &\bf{0.1821}    \\
			\hline
		\end{tabular}
		\label{Comparison between RDFE and FFN}
	\end{center}
\end{table}

\begin{table}[t]
	\begin{center}
		\caption{Ablation study of our RDFE  \textcolor{black}{(CelebA, $\times 8$)}.}
        \setlength{\tabcolsep}{2.5mm}
        \renewcommand\arraystretch{1}
		\begin{tabular}{c|cccc}
			\hline
			Methods      & PSNR${\uparrow}$  & SSIM${\uparrow}$  & VIF${\uparrow}$  & LPIPS${\downarrow}$      \\
			\cline{2-5}
			\hline
			\hline
			Single path (${3\times3}$ dw)  & 27.73     &0.7934       & 0.4619      &0.1915 \\
            Single path (${5\times5}$ dw)  & 27.71     & 0.7912     &0.4587       &0.1944 \\
            Single path (${7\times7}$ dw)  &27.72     &0.7926      &0.4602       &0.1922 \\
			RDFE w/o AU      & 27.76      &0.7951       &0.4673        &0.1846 \\
            RDFE w/o FRM    & 27.75      &0.7941       & 0.4643       &0.1928  \\
		   RDFE             &\bf{27.83}  &\bf{0.7961}  &\bf{0.4725}   &\bf{0.1821}  \\
			\hline
		\end{tabular}
		\label{Effects of RDFEM}
	\end{center}
\end{table}

\begin{table}[t]
 \begin{center}
  \caption{Ablation study of our SKAF \textcolor{black}{(CelebA, $\times 8$).}}
  \setlength{\tabcolsep}{2mm}
        \renewcommand\arraystretch{1}
  \begin{tabular}{cccc|cc}
  \hline
      $5\times5$ conv  &$7\times7$ conv  &Avepool  &Maxpool &PSNR${\uparrow}$  &SSIM${\uparrow}$ \\
        \hline
        \hline
         ${\times}$  & ${\times}$   & ${\times}$ &${\times}$ &27.69 &0.7919\\
        $\times$   & ${\surd}$  & $\surd$  & $\surd$  & 27.76 &0.7955  \\ 
        ${\surd}$   & ${\times}$  & $\surd$  & $\surd$  &27.73 &0.7946  \\
        ${\times}$  & $\surd$     & ${\times}$ & ${\times}$ & 27.74 &0.7931  \\ 
        ${\times}$   & $\surd$ & $\surd$     & ${\times}$  & 27.79 &0.7951 \\  
        ${\times}$   & $\surd$    & ${\times}$  & $\surd$    & 27.78 &0.7946 \\  
        ${\surd}$  & $\surd$     & $\surd$  & $\surd$    & \bf{27.83}  &\bf{0.7961}\\   
        \hline
  \end{tabular}
  \label{Effects of SKAFM}
 \end{center}
\end{table}

\begin{table}[t]
	\begin{center}
		\caption{Ablation study of our EDFF  \textcolor{black}{(CelebA, $\times 8$)}.}
        \setlength{\tabcolsep}{1mm}
        \renewcommand\arraystretch{1}
		\begin{tabular}{c|ccc|ccc}
			\hline
            \multirow{2}{*}{Methods} & \multicolumn{3}{c|}{Baseline} & \multicolumn{3}{c}{+ Our EDFF} \\
			\cline{2-7}
			& Parameters & PSNR${\uparrow}$  & SSIM${\uparrow}$ & Parameters & PSNR${\uparrow}$  & SSIM${\uparrow}$      \\
			\cline{2-5}
			\hline
			\hline
			SPARNet~\cite{chen2020learning}          &10.6M       &27.73       &0.7949       &12.1M &\textbf{27.85} &\textbf{0.7964} \\
		  SFMNet~\cite{wang2023spatial}     &8.6M       &27.96      &0.7996    &10.7M &\textbf{28.07} &\textbf{0.8011} \\
		   
			\hline
		\end{tabular}
		\label{Effects of EDFFM}
	\end{center}
\end{table}

\section{Experiments}
\label{sec4}

\subsection{Datasets and Evaluation Metrics}
\label{sec41}
We utilize the CelebA~\cite{liu2015deep} dataset for training and evaluation on CelebA~\cite{liu2015deep}, Helen~\cite{le2012interactive}, and SCface~\cite{grgic2011scface} datasets, respectively. We center-crop the aligned faces and resize them to ${128\times128}$ pixels to obtain HR images. These HR images are then downsampled to ${16\times16}$ pixels using bicubic interpolation, producing the corresponding LR images. In our experiments, we randomly chose 18,000 CelebA images for training and 1,000 for testing. In addition, we also utilize the SCface test set as a real evaluation dataset. To measure the quality of FSR results, we use PSNR~\cite{wang2004image}, SSIM~\cite{wang2004image}, LPIPS~\cite{zhang2018unreasonable}, VIF~\cite{sheikh2006image}, and FID~\cite{obukhov2020quality}.

\subsection{Implementation details}
\label{sec42}
We implement our model using PyTorch on an NVIDIA GeForce RTX 3090. The network is optimized using the Adam optimizer, with parameters set to ${\beta_1 = 0.9}$ and ${\beta_2 = 0.99}$. The initial learning rate is ${2 \times 10^{-4}}$, with separate learning rates for the generator and discriminator set at ${1 \times 10^{-4}}$ and ${4 \times 10^{-4}}$, respectively. The loss function weights are configured as ${\lambda_{pix} = 1}$, ${\lambda_{pcp} = 0.01}$, and ${\lambda_{adv} = 0.01}$.

\begin{table*}[t]
    \caption{Quantitative comparisons of ours and existing FSR methods for $\times 8$ FSR on CelebA and Helen test sets.}
	\captionsetup{aboveskip=5pt}
	\captionsetup{belowskip=-10pt}
	\small
	\centering
	\scalebox{1}{
		\begin{tabular}{p{2.3cm}|p{1.4cm}p{1.4cm}p{1.4cm}p{1.4cm}|p{1.4cm}p{1.4cm}p{1.4cm}p{1.4cm}p{1.4cm}}
			\toprule
			\multirow{2}{*}{Methods}  & \multicolumn{4}{c|}{CelebA~\cite{liu2015deep}}   &\multicolumn{4}{c}{Helen~\cite{le2012interactive}}\\
			\cline{2-5} \cline{6-9} 
			
			& PSNR${\uparrow}$ & SSIM${\uparrow}$ & VIF${\uparrow}$ & LPIPS${\downarrow}$ &PSNR${\uparrow}$  & SSIM${\uparrow}$ & VIF${\uparrow}$ & LPIPS${\downarrow}$  \\
			\hline
			\hline
			Bicubic                           & 23.61 & 0.6779 & 0.1821 & 0.4899 & 22.95 & 0.6762 & 0.1745 & 0.4912  \\
			
			SAN~\cite{dai2019second}          & 27.43 & 0.7826 & 0.4553 & 0.2080 & 25.46 & 0.7360 & 0.4029 & 0.3260  \\
			
			RCAN~\cite{zhang2018image}        & 27.45 & 0.7824 & 0.4618 & 0.2205 & 25.50 & 0.7383 & 0.4049 & 0.3437  \\
			
			HAN~\cite{niu2020single}          & 27.47 & 0.7838 & 0.4673 & 0.2087 & 25.40 & 0.7347 & 0.4074 & 0.3274  \\
			
			SwinIR~\cite{liang2021swinir}     & 27.88 & 0.7967 & 0.4590 & 0.2001 & 26.53 & 0.7856 & 0.4398 & 0.2644 \\
			
			FSRNet~\cite{chen2018fsrnet}      & 27.05 & 0.7714 & 0.3852 & 0.2127 & 25.45 & 0.7364 & 0.3482 & 0.3090  \\
			
			DICNet~\cite{ma2020deep}          & 27.42     & 0.7840      & 0.4234      & 0.2129     & 26.15 & 0.7717 & 0.4085 & \underline{0.2158} \\
			
			FACN~\cite{xin2020facial}         & 27.22 & 0.7802 & 0.4366 & \underline{0.1828} & 25.06 & 0.7189 & 0.3702 & 0.3113  \\
			
			SPARNet~\cite{chen2020learning}   & 27.73 & 0.7949 & 0.4505 & 0.1995 & 26.43 & 0.7839 & 0.4262 & 0.2674  \\
			
			SISN~\cite{lu2021face}            & 27.91 & 0.7971 & \underline{0.4785} & 0.2005 & 26.64 & 0.7908 & 0.4623 & 0.2571  \\
 
	        AD-GNN~\cite{bao2022attention}            & 27.82 & 0.7962 & 0.4470
	        & 0.1937 & 26.57 & 0.7886 & 0.4363 & 0.2432  \\
	        
	    Restormer-M~\cite{zamir2022restormer}            & 27.94 & \underline{0.8027} & 0.4624 & 0.1933 & \underline{26.91} & \underline{0.8013} & \underline{0.4595} & 0.2258  \\
	        
	        LAAT~\cite{li2023learning}            & 27.91 & 0.7994 & 0.4624 & 0.1879 & 26.89 & 0.8005 & 0.4569 & 0.2255  \\
	        ELSFace~\cite{10145603}  &27.41 &0.7922 &0.4451 &0.1867 &26.04 &0.7873 &0.4193 &0.2811 \\
	        SFMNet~\cite{wang2023spatial}            & \underline{27.96} & 0.7996 & 0.4644 & 0.1937 & 26.86 & 0.7987 & 0.4573 & 0.2322  \\
	        SPADNet~\cite{wang2024structure}       &27.82      &0.7966      &0.4589      &0.1987    &26.47   &0.7857    &0.4295    &0.2654  \\
		    \midrule
		AMINet  &{\bf28.26} &{\bf0.8091} &{\bf0.4893} &{\bf0.1755} &{\bf27.01} &{\bf0.8042} &{\bf0.4694} &{\bf0.2067} \\
				
	    \bottomrule
		\end{tabular}
	}
	\label{compare_CelebA_Helen}
\end{table*}

\begin{figure*}[t]
	\centerline{\includegraphics[width=17.8cm]{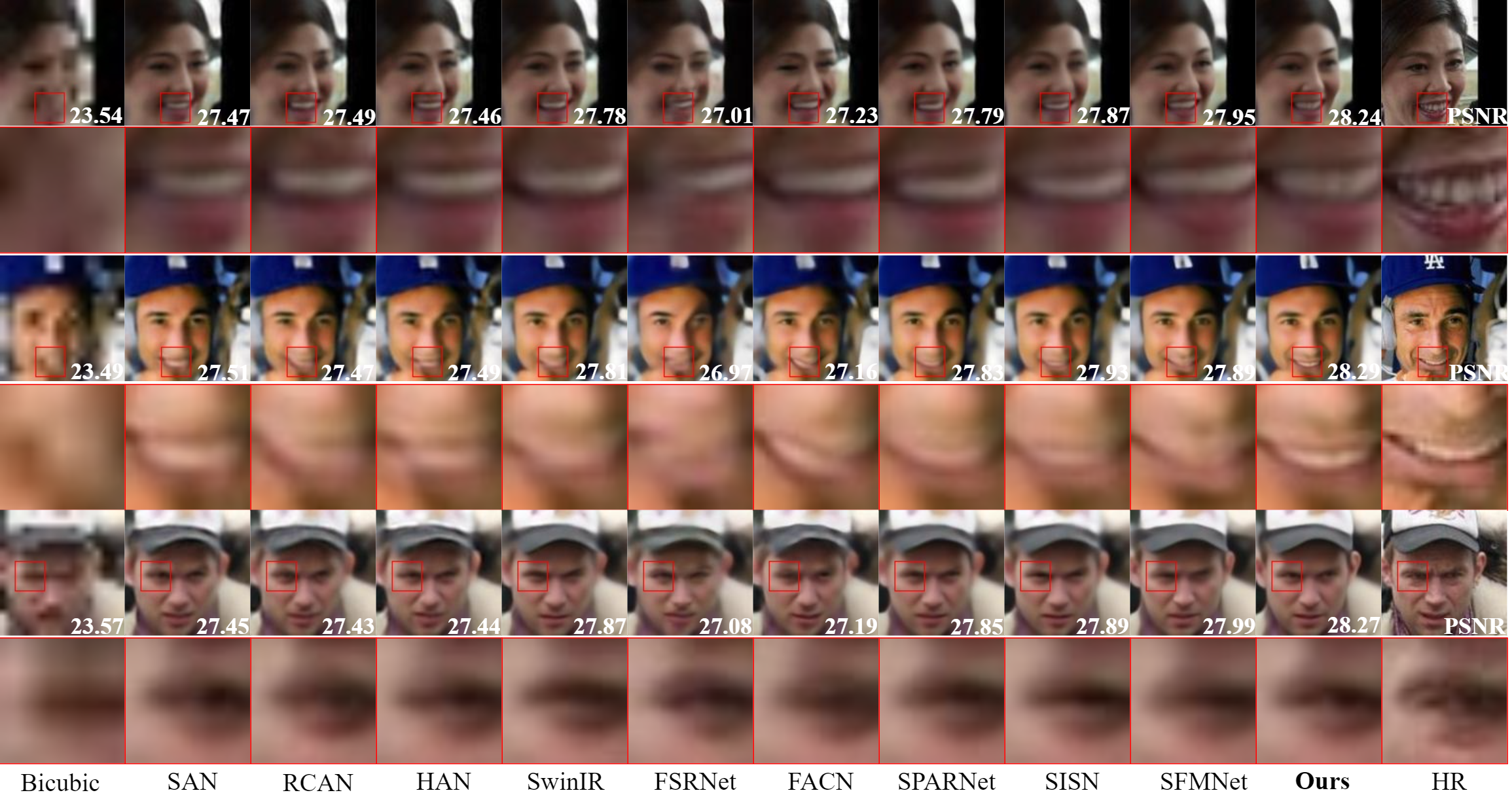}}
	\caption{Visual comparisons for $\times$8 FSR on CelebA~\cite{liu2015deep} test set. Our method can recover accurate face images.}
	\label{compare_CelebA}
\end{figure*}

\subsection{Ablation Studies}
\label{sec43}

\subsubsection{Study of LGFI} 
LGFI is proposed to extract local features and global relationships of images, which represents a new attempt to interact with local and global information. To verify the reasonableness of our design of LGFI, as shown in Table~\ref{Effects of LGFIM}, we design four ablation models. The first model removes the SA, labeled “LGFI w/o SA”. The second model removes RDFE, labeled as “LGFI w/o RDFE”. The third model removes SKAF, labeled as “LGFI w/o SKAF”. We have the following observations: (a) Introducing SA and RDFE alone can improve model performance. This is because the above two modules can capture local and global features to promote facial feature reconstruction, including facial details and overall contours. (b) Model performance has been significantly increased by introducing the SKAF to capture the relationship between local and global facial features. This is because our SKAF can promote interaction between our SA and RDFE, integrating richer information and providing supplementary information for the final FSR.

Furthermore, we provide a visual comparison in Fig.~\ref{compare_lgfi}, illustrating the impact of removing certain components from LGFI. The reconstructed images exhibit noticeable blurring or artifacts, highlighting the importance of these components. Additionally, Fig.~\ref{heatmap_lgfi} presents heatmap visualizations of the outputs from different components within LGFI, with their corresponding locations in the network indicated in Fig.~\ref{Network structure}. Specifically, SKAF effectively integrates the global contours captured by SA with the regional features extracted by RDEF. This integration enables the model to focus more on essential facial structures and components while reducing emphasis on less critical details, such as hair. 
In addition, we quantitatively evaluate the computational efficiency of each module by comparing inference latency and parameter counts in Table~\ref{Effects of LGFIM}, where the RDEF module has the greatest impact on both inference time and parameter counts. This is attributed to its multi-branch fusion strategy and deep refinement operations, which introduce additional computational complexity. However, RDEF delivers substantial performance improvements, which is overall worthwhile.

\begin{figure*}[t]
	\centerline{\includegraphics[width=17.8cm]{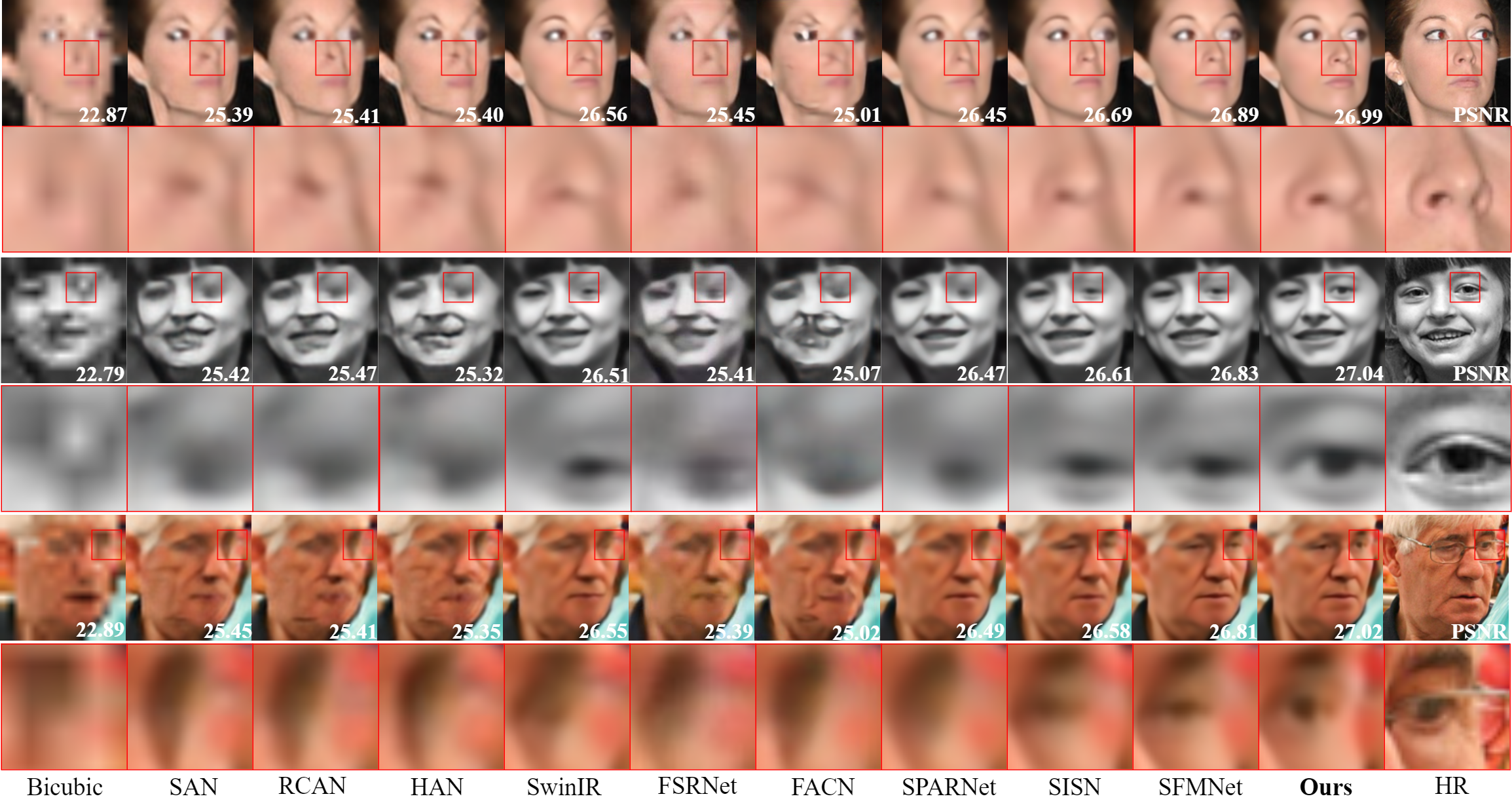}}
	\caption{Visual comparisons for $\times$8 FSR on Helen~\cite{le2012interactive} test set. Our method can recover accurate face images.}
	\label{compare_Helen}
\end{figure*}

\subsubsection{Comparison between LGFI and Transformer}
As shown in Fig.~\ref{fig:compare_structure}, LGFI uses a dual-branch structure to represent the local and global features. In contrast, the traditional Transformer in Restormer~\cite{zamir2022restormer} uses a serial structure to link local and global features. To verify the effectiveness of LGFI, we replace all LGFIs with Transformers and conduct comparative experiments with similar parameters between the two models. From Table~\ref{Comparison between LGfim and Trans}, the network's performance using LGFI is better when the two networks maintain similar parameters. This is because LGFI utilizes the features of both local and global branches for interaction, facilitating the communication of multiscale facial information.

\subsubsection{Comparison between RDFE and FFN}
The feed-forward network (FFN) performs independent nonlinear transformations of the inputs at each position to help the Transformer capture local features, but it cannot extract multiscale features, which is not favorable for accurate FSR. In contrast, our RDFE can extract multiscale local features well. To compare RDFE and FFN, we replace RDFE with FFN while keeping the parameters of the two models similar. As shown in Table~\ref{Comparison between RDFE and FFN}, since FFN's ability to capture feature interactions is limited compared to our RDFE, which utilizes multiple branches to capture different receptive field features, our RDFE performs better than FFN with similar computational cost.

\subsubsection{Effectiveness of RDFE}
In RDFE, a three-branch network guided by an attention mechanism is used for deep feature extraction, and the feature refinement module is used to enrich feature representation. To verify the effectiveness of RDFE, we conduct multiple ablation experiments. We designed five improved models. The first model adopts a single branch structure of ${3 \times3} $ depthwise convolution, labeled as “Single path (${3 \times3}$ dw)”. The second model adopts a single branch structure of ${5 \times5} $ depthwise convolution, labeled as “Single path (${5 \times5}$ dw)”. The third model adopts a single branch structure of ${7 \times7} $ depthwise convolution, labeled as “Single path (${7 \times7}$ dw)”. The fourth model removes attention units labeled as "w/o AU". The fifth model removes the feature refinement module, labeled as "w/o FRM". From the Table~\ref{Effects of RDFEM}, we have the following observations:
(a) By comparing the first three rows and the last row of the table, it can be seen that multiscale branching facilitates the model's performance due to its ability to extract face features at different levels;
(b) From the comparison between the second and the last rows of the table and the last row, it can be seen that using attention units (AU) to guide three-branch feature extraction can enable the model to adaptively allocate weights, enhance the representation of important facial information, and thus improve model performance;
(c) From the last two rows of the table, we can conclude that the feature refinement module (FRM) can further integrate multiscale information, refine multiscale fusion features, and thus improve performance.

\subsubsection{Effectiveness of SKAF} SKAF is an important component of LGFI, facilitating information exchange between local and global branches. We perform ablation experiments to validate our SKAF module's impact and assess the combined approach's practicality. Since SKAF consists of dual-branch convolutional layers, maximum pooling layers, and average pooling layers, we verify the effectiveness of module components in SKAF. From Table~\ref{Effects of SKAFM}, we have the following observations: (a) From the last three rows of the table, we find that using a single pooling branch results in reduced performance, while using average pooling alone results in lower performance than using maximum pooling alone. This is because the salient features of the face are the key to facial recovery, with maximum pooling focusing on salient facial feature information. In contrast, average pooling focuses on the overall information of the face. (b) Compared to the third and fifth rows, it can be concluded that using both ${5 \times5} $ and ${7 \times7} $ simultaneously can improve performance and fully utilize key facial information under different receptive fields.

\subsubsection{Study of EDFF}
This section presents a set of experiments to validate the effectiveness of our EDFF, a module tailored for fusing multiscale features. We add EDFF to SPARNet~\cite{chen2020learning}, which uses EDFF to connect the encoding and decoding stages in SPARNet and send them to the next decoding stage. Additionally, we add EDFF to SFMNet~\cite{wang2023spatial}, and the specific operation is the same as in SPARNet. From the results of Table~\ref{Effects of EDFFM}, we can see that although the parameters of both models increase slightly, the performance of the models improves, which precisely proves that EDFF is helpful for feature fusion in the encoding and decoding stages.

\begin{figure}[t]
\centering
\includegraphics[width=8.5cm, trim=10 0 0 10]{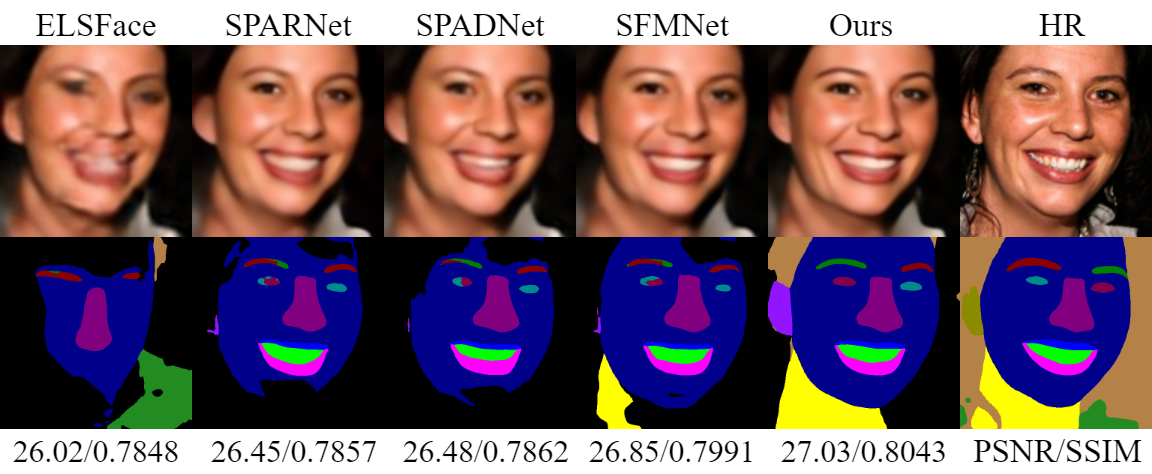}
\caption{Comparisons of face parsing on the Helen test set.} 
\label{helen_parsing}
\vspace{-4mm}
\end{figure}

\begin{figure}[t]
\centering
\includegraphics[width=8.5cm, trim=10 0 0 10]{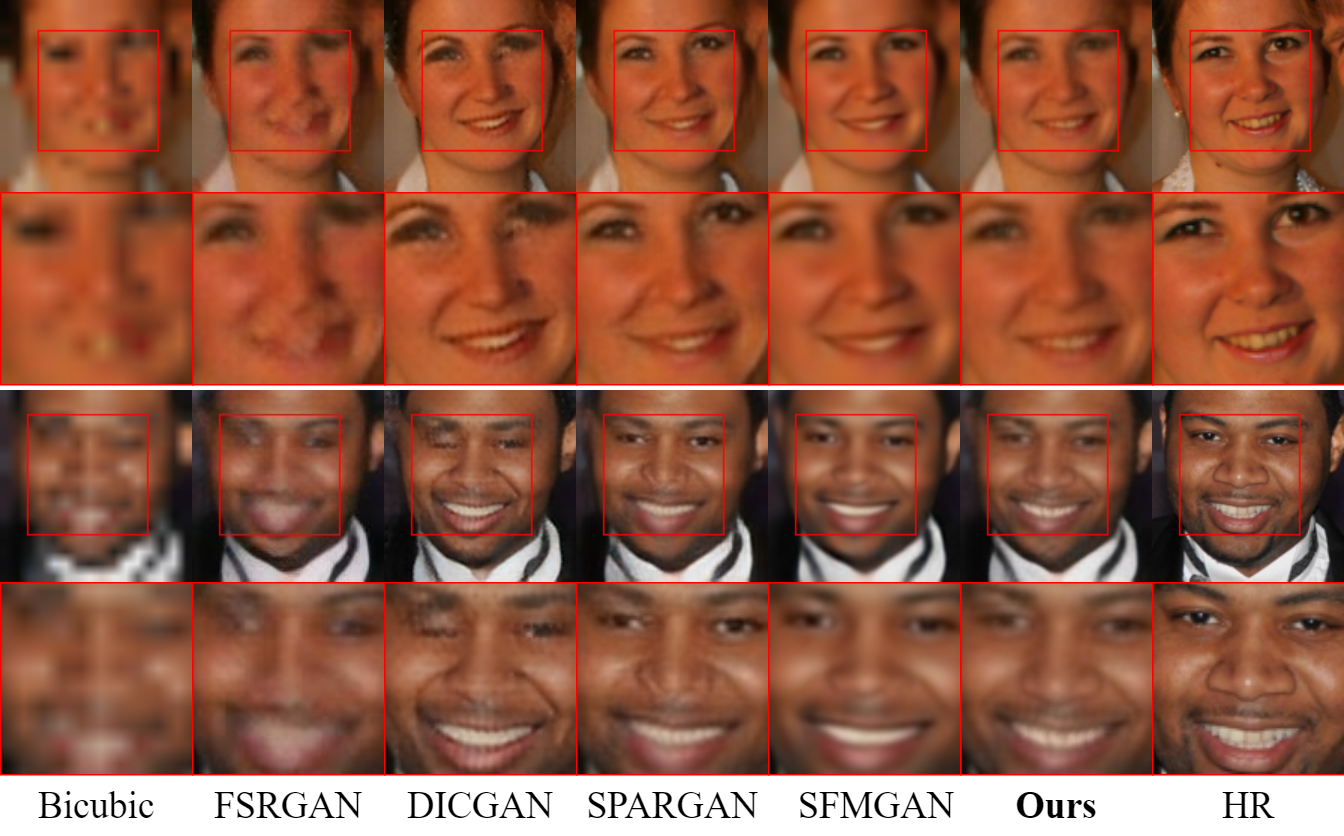}
\caption{Visual comparisons of existing GAN-based FSR methods on the Helen test set. Our AMIGAN can reconstruct high-quality face images with clear facial components.}
\label{GAN Comparision}
\end{figure}

\begin{table}[t]
	\begin{center}
		\caption{Quantitative comparison of ours with other GAN-based methods (Helen, $\times 8$).}
        \setlength{\tabcolsep}{3.2mm}
        \renewcommand\arraystretch{1.2}
		\begin{tabular}{c|cccc}
			\hline
			Methods  & PSNR${\uparrow}$  & SSIM${\uparrow}$  & VIF${\uparrow}$  & FID${\downarrow}$      \\
			\hline
			\hline
                FSRGAN~\cite{chen2018fsrnet} & 25.02  & 0.7279    & 0.3400    & 146.55  \\
                DICGAN~\cite{ma2020deep}  & 25.59  & 0.7398   & 0.3925  & 144.25 \\
                SPARGAN~\cite{chen2020learning} & 25.86   & 0.7518    & 0.3932  & 149.54 \\
             SFMGAN~\cite{wang2023spatial}  & 25.96       & 0.7618       & 0.4019       & 141.23 \\
            AMIGAN (Ours)     & \textbf{26.35}  & \textbf{0.7769} & \textbf{0.4101}  & \textbf{122.43} \\
			\hline
		\end{tabular}
		\label{comparision with gan}
	\end{center}
\end{table}

\begin{table}[t]
	\begin{center}
		\caption{Comparisons of cosine similarity on $\times 8$ SCface. }
        \setlength{\tabcolsep}{3mm}
        \renewcommand\arraystretch{1.0}
		\begin{tabular}{c|c c c c}
			\hline
            \multirow{2}{*}{Methods} & \multicolumn{4}{c}{Cosine Similarity${\uparrow}$}  \\  
            \cline{2-5}
                                      & Case 1 & Case 2 & Case 3 & Case 4 \\
            \hline 
			\hline
	        SAN~\cite{dai2019second}        & 0.8133 & 0.8145 & 0.8244 &0.8192  \\		
	        RCAN~\cite{zhang2018image}      & 0.8196 & 0.8214 & 0.8199 &0.8201  \\	
	        FSRNet~\cite{chen2018fsrnet}    & 0.8032 & 0.7982 & 0.8105 &0.8087  \\	
            FACN~\cite{xin2020facial}   &0.8002    & 0.7989    &0.8115      &0.8014   \\
	        SPARNet~\cite{chen2020learning} & 0.8215 & 0.8209 & 0.8244 &0.8195  \\		
            SISN~\cite{lu2021face}          & 0.8345 & 0.8378 & 0.8373 &0.8391  \\
             LAAT~\cite{li2023learning}   & \underline{0.8501}  & \underline{0.8497}    & \underline{0.8456}      & \underline{0.8472}    \\
            ELSFace~\cite{10145603}         & 0.8165 & 0.8124 & 0.8136 &0.8148  \\
            SFMNet~\cite{wang2023spatial}   & 0.8411 & 0.8456 & 0.8429 &0.8402  \\
            \midrule
	        AMINet      & {\bf 0.8533} & {\bf 0.8577} &{\bf 0.8601} & {\bf0.8501 }  \\
			\hline
		\end{tabular}
		\label{SCface}
	\end{center}
\end{table}

\begin{figure*}[h]
\centerline{\includegraphics[width=17.8cm]{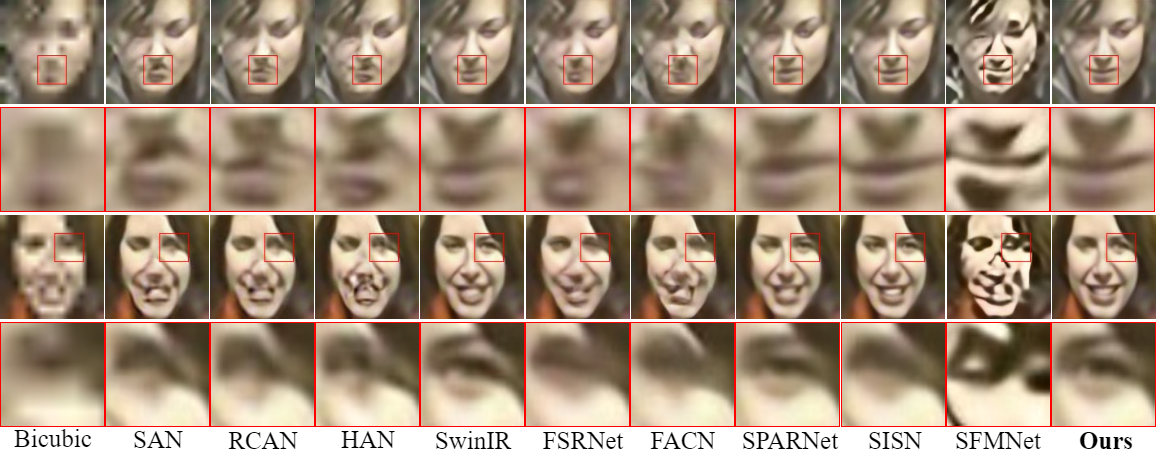}}
\caption{Visual comparisons for $\times$8 FSR on SCface~\cite{grgic2011scface} test set. Our method recovers clearer face images.}
\label{compare_SCface}
\end{figure*}

\begin{table}[t]
	\begin{center}
		\caption{Comparisons of model complexity on $\times 8$ CelebA.}
        \setlength{\tabcolsep}{3.2mm}
        \renewcommand\arraystretch{1.0}
		\begin{tabular}{c|cccc}
			\hline
			Methods    & PSNR${\uparrow}$   & Params${\downarrow}$ & MACs${\downarrow}$ & Speed${\downarrow}$         \\
			\hline
			\hline

	FSRNet~\cite{chen2018fsrnet}   &27.05   & 27.5M & 40.7G  & 89ms  \\	
	SPARNet~\cite{chen2020learning} &27.73  & 16.6M & 7.1G & 40ms \\	 
   DICNet~\cite{ma2020deep}    &27.42    & 22.8M  & 35.5G &121ms  \\  
    AD-GNN~\cite{bao2022attention}   &27.82    &15.8M & 15.0G &108ms \\

   CTCNet~\cite{gao2023ctcnet}  &28.37  &22.4M &47.2G &106ms    \\ 
   SCTANet~\cite{bao2023sctanet}        & 28.26  &27.7M &10.4G &58ms \\     
    LAAT~\cite{li2023learning}            & 27.91 &22.4M  &8.9G    &36ms\\
    ELSFace~\cite{10145603}   &27.41 &6.8M &10.6G &32ms    \\
     SFMNet~\cite{wang2023spatial}      & 27.96     & 8.6M  &30.6G  & 48ms   \\
               \midrule
	AMINet-S  &27.93    & 8.2M    &9.4G     &25ms  \\
    AMINet   &{28.26} &{12.62M} &{15.6G} &{35ms}   \\
			\hline
		\end{tabular}
		\label{model analysis}
	\end{center}
\end{table}

\subsection{Comparisons with Other Methods}
\label{sec44}

We compares AMINet with existing FSR methods, including SAN~\cite{dai2019second}, RCAN~\cite{zhang2018image}, HAN~\cite{niu2020single}, SwinIR~\cite{liang2021swinir}, FSRNet~\cite{chen2018fsrnet}, DICNet~\cite{ma2020deep}, FACN~\cite{xin2020facial}, SPARNet~\cite{chen2020learning}, SISN~\cite{lu2021face},  AD-GNN~\cite{bao2022attention}, Restormer-M~\cite{zamir2022restormer}, LAAT~\cite{li2023learning}, ELSFace~\cite{10145603}, SFMNet~\cite{wang2023spatial} and SPADNet~\cite{wang2024structure}.

\subsubsection{Comparisons on CelebA dataset} We conduct a quantitative comparison of AMINet against existing FSR methods on the CelebA test set, as detailed in Table~\ref{compare_CelebA_Helen}. Our AMINet outperforms all evaluation metrics, including PSNR, SSIM, LPIPS, and VIF, which fully demonstrates its efficiency. This strongly validates the effectiveness of AMINet. Additionally, the visual comparison in Fig.~\ref{compare_CelebA} reveals that previous FSR methods struggled to reproduce facial features like the eyes and mouth accurately. In contrast, AMINet excels at preserving the facial structure and producing more precise results.

\subsubsection{Comparisons on Helen dataset} 

We evaluate our method on the Helen test set to further assess AMINet's versatility. Table~\ref{compare_CelebA_Helen} provides a quantitative comparison of $\times$8 FSR results about it, where AMINet achieves the better performance. Visual comparisons in Fig.~\ref{compare_Helen} indicate that existing FSR methods struggle to maintain accuracy, leading to blurred shapes and a loss of facial details. In contrast, AMINet successfully preserves facial contours and details, reinforcing its effectiveness and adaptability across different datasets. In addition, we provide a visual comparison of face parsing maps for recovered face images, as shown in Fig.~\ref{helen_parsing}, which clearly shows that our AMINet facilitates downstream tasks such as face parsing maps segmentation.

\subsubsection{Comparisons with GAN-based methods}
We present AMIGAN as a novel approach to enhance the visual fidelity of image restoration. To demonstrate its effectiveness, we compare AMIGAN with existing GAN-based methods, including FSRGAN~\cite{chen2018fsrnet}, DICGAN~\cite{ma2020deep}, SPARGAN~\cite{chen2020learning}, and SFMGAN~\cite{wang2023spatial}. In addition to conventional metrics, we adopt FID~\cite{obukhov2020quality} for quantitative evaluation. Results on the Helen dataset (Table~\ref{comparision with gan}) show that AMIGAN consistently outperforms prior methods. Visual comparisons in Fig.~\ref{GAN Comparision} further highlight AMIGAN’s superior ability to restore fine facial structures and texture details, particularly around the mouth and nose, delivering clearer and more realistic reconstructions with fewer artifacts.

\subsubsection{Comparisons on Real-world surveillance faces} 
All the above comparisons are tested on synthetic test sets, which fail to simulate real-world scenarios accurately. To further evaluate our model's performance in real-world conditions, we also conduct experiments using low-quality face images from the SCface dataset~\cite{grgic2011scface}. As shown in Fig.~\ref{compare_SCface}, we compare the reconstruction results. From this figure, we find that the reconstruction results of face prior-based methods are not satisfactory. The challenge lies in accurately estimating priors from real-world LR facial images. Incorrect prior information can lead to misleading guidance during the reconstruction process. In contrast, our AMINet can restore clearer face details and faithful face structures. As shown in Table~\ref{SCface}, we also provide a quantitative comparison of cosine similarity using the above methods. This result fully demonstrates our method's effectiveness in real scenarios.

\subsection{Model Complexity and Convergence Analysis}
\label{sec45}
In addition to the performance indicators mentioned earlier, the number of model parameters, inference time, and computational complexity are crucial factors in evaluating performance. As shown in the Table~\ref{model analysis}, we have selected some models for comparison. Meanwhile, as shown in Fig.~\ref{fig: Model_Complexity}, we compare our model with existing ones in terms of parameters, PSNR values, and inference speed. We can see that AMINet can have faster inference time, smaller parameter count, and computational complexity with similar performance to SCTANet. In addition, we provide a small parameter version called "AMINet-S". AMINet-S performs similarly to SFMNet in terms of parameter quantity, while AMINet has more advantages in computational complexity and inference time. This is thanks to AMINet achieving high efficiency by promoting extensive multiscale feature exchange within the network. This design enables features from different receptive fields to interact effectively, allowing the model to select the most relevant information for facial reconstruction adaptively. Therefore, AMINet maintains strong FSR performance while balancing parameters and computational costs.

\begin{figure}[t]
\centering
\includegraphics[width=9cm, trim=23 10 23 10, clip]{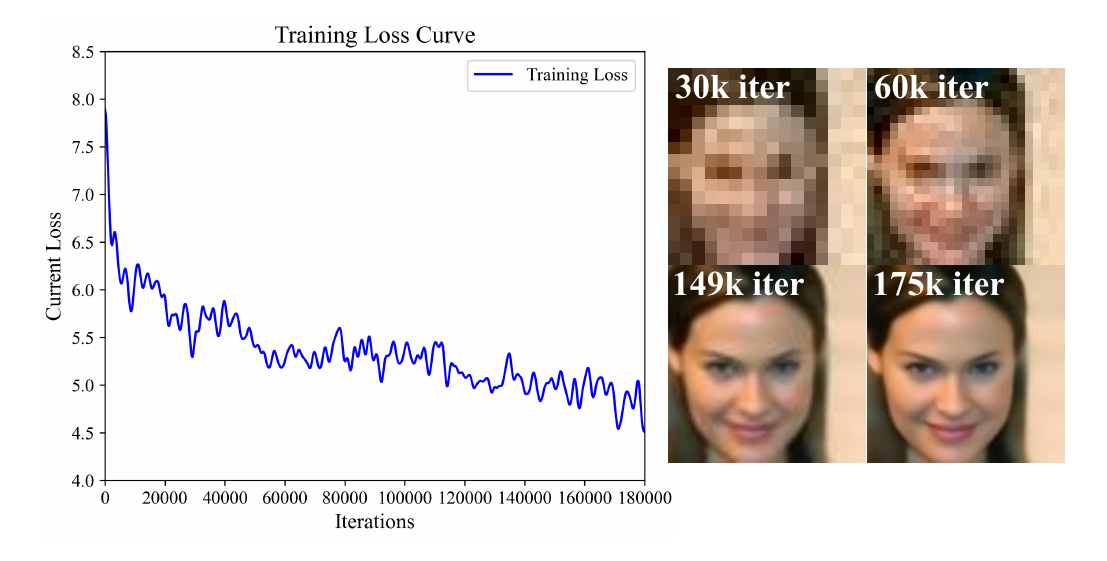}
\caption{Model convergence and visualization analysis.}
\label{loss}
\end{figure}

We also provide the training loss curve in Fig.~\ref{loss}, which shows a consistent decrease as iterations progress, indicating stable optimization and convergence. For clarity, the loss values are scaled by a factor of 100. We also visualize the intermediate results at different training stages, illustrating the clear improvement from early to late iterations as the model progressively learns more discriminative features. These observations collectively confirm the effectiveness of the training strategy and its stable convergence properties.

\section{Discussion and Future Works} \label{sec5}
Although our AMINet performs well in FSR, it still has certain limitations. The model's robustness to extreme poses at very low resolutions, such as side profiles, and its ability to handle challenging lighting conditions, including low-light and overexposed environments, require further improvement. Moreover, while our method significantly reduces computational cost compared to existing methods, it remains insufficient for deployment on mobile devices.

Future work will focus on enhancing AMINet’s robustness to extreme poses and challenging lighting conditions while accelerating inference. This includes integrating pose-invariant feature learning through attention-based mechanisms or 3D-aware priors for better side-profile restoration and developing adaptive illumination correction using physics-based relighting models or low-light enhancement. Additionally, optimizing the model with lightweight architectures, quantization, and efficient inference strategies will enable faster inference while maintaining high-quality restoration.

\section{Conclusions}
\label{sec6}

We propose an attention-guided multiscale interaction network for face super-resolution. The core component, LGFI, facilitates effective interaction between global features from self-attention and local features extracted by the proposed RDFE module. To enrich local representations, RDFE employs multiscale depthwise separable convolutions combined with attention for feature extraction and refinement. Moreover, an adaptive kernel selection mechanism further promotes multiscale feature fusion. Extensive experiments on synthetic and real-world datasets demonstrate that our design substantially enhances cross-scale feature interaction, enabling our method to surpass existing approaches in reconstruction quality, model size, and inference efficiency.

\bibliographystyle{IEEEtran}
\bibliography{reference}

%

\begin{IEEEbiography}[{\includegraphics[width=1in,height=1.25in,clip,keepaspectratio]{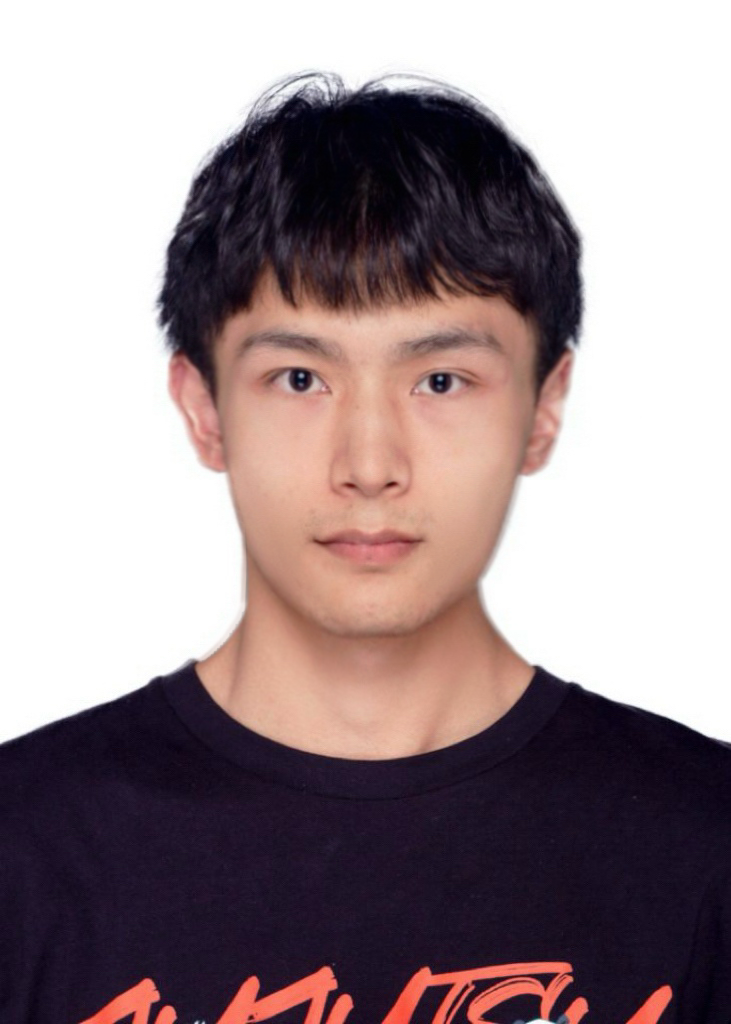}}]{Xuejie Wan}
received the B.S. degree in Automation from the School of Microelectronics and Control Engineering, Changzhou University, Changzhou, China, in 2022, and the M.S. degree in Control Science and Engineering from the College of Automation, Nanjing University of Posts and Telecommunications, Nanjing, China, in 2025. His research interest includes image super-resolution.
\end{IEEEbiography}

\begin{IEEEbiography}[{\includegraphics[width=1in,height=1.25in,clip,keepaspectratio]{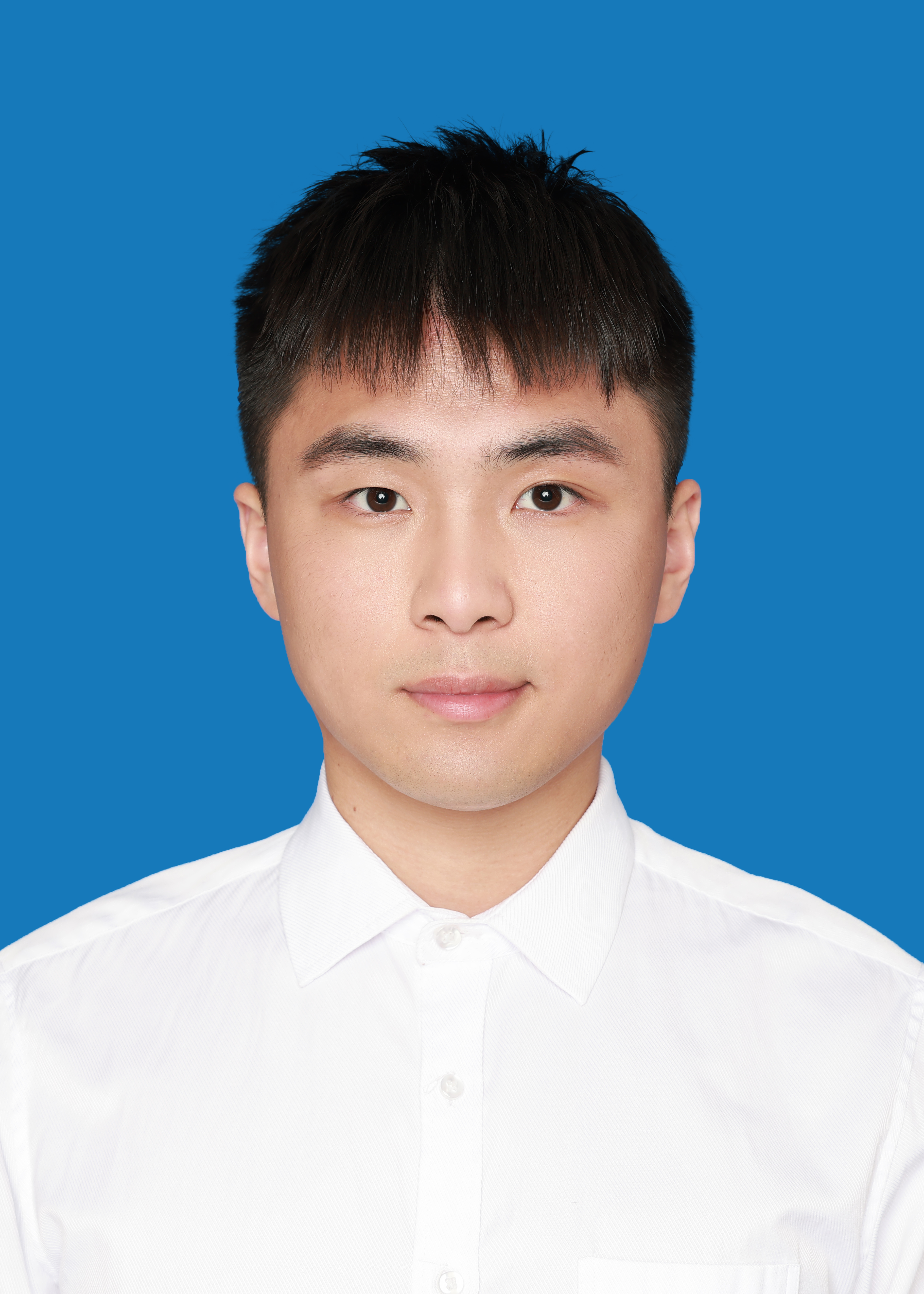}}]{Wenjie Li}
received the M.S. degree in control science and engineering from the College of Automation, Nanjing University of Posts and Telecommunications, Nanjing, in 2023. He is currently pursuing the Ph.D. degree in artificial intelligence with the School of Artificial Intelligence, Beijing University of Posts and Telecommunications. His research interests include image restoration.
\end{IEEEbiography}

\begin{IEEEbiography}[{\includegraphics[width=1in,height=1.25in,clip,keepaspectratio]{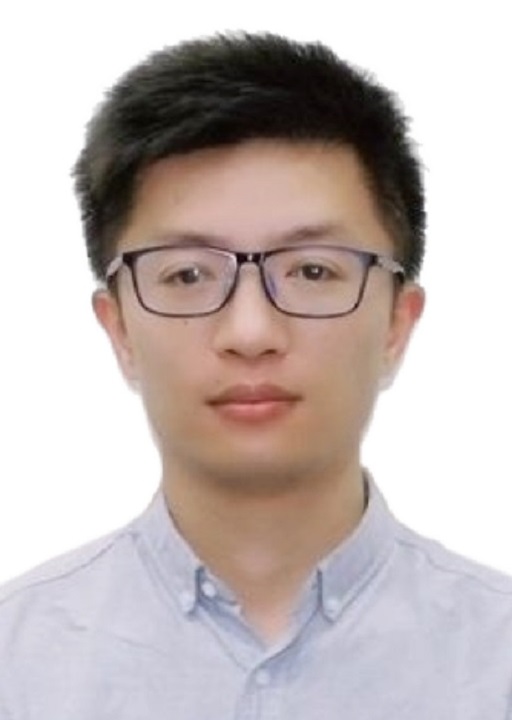}}]{Guangwei Gao}
(Senior Member, IEEE) received the Ph.D. degree in Pattern Recognition and Intelligent Systems from the Nanjing University of Science and Technology, Nanjing, in 2014. He was a visiting student of the Department of Computing, The Hong Kong Polytechnic University, in 2011 and 2013, respectively. From 2019 to 2021, he was a Project Researcher with the National Institute of Informatics, Japan. He is currently a Professor at Nanjing University of Posts and Telecommunications. His research interests include pattern recognition and computer vision. Personal website: \textit{https://guangweigao.github.io}.
\end{IEEEbiography}

\begin{IEEEbiography}[{\includegraphics[width=1in,height=1.25in,clip,keepaspectratio]{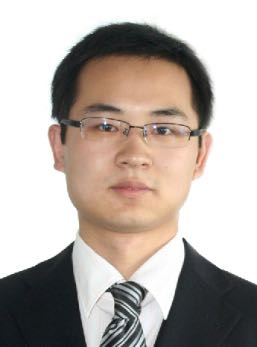}}]{Huimin Lu}
(Senior Member, IEEE) received the Ph.D. degree in Electrical Engineering from the Kyushu Institute of Technology, Kitakyushu, Japan, in 2014. From 2013 to 2016, he was a JSPS Research Fellow with the Kyushu Institute of Technology. From 2016 to 2024, he was an Associate Professor with the Kyushu Institute of Technology and an Excellent Young Researcher of Ministry of Education, Culture, Sports, Science and Technology. He is currently a Professor with Southeast University, Nanjing, China. His research interests include artificial intelligence, computer vision, and robotics.
\end{IEEEbiography}

\begin{IEEEbiography}[{\includegraphics[width=1in,height=1.25in,clip,keepaspectratio]{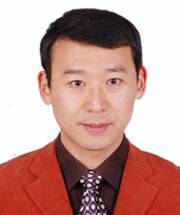}}]{Jian Yang}
(Member, IEEE) received the Ph.D. degree in pattern recognition and intelligent systems from the Nanjing University of Science and Technology (NJUST), Nanjing, China, in 2002. From 2006 to 2007, he was a postdoctoral fellow with the Department of Computer Science, New Jersey Institute of Technology. He is currently a professor with the School of Computer Science and Technology, NJUST. He is the author of more than 400 scientific papers in pattern recognition and computer vision. 
His research interests include pattern recognition, computer vision, and machine learning. He is/was an associate editor for Pattern Recognition and IEEE Transactions on Neural Networks and Learning Systems. He is a fellow of IAPR.
\end{IEEEbiography}

\begin{IEEEbiography}[{\includegraphics[width=1in,height=1.25in,clip,keepaspectratio]{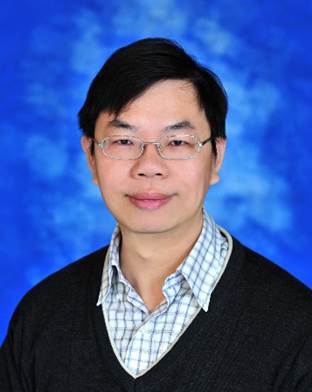}}]{Chia-Wen Lin}
(Fellow, IEEE) received the Ph.D. degree in Electrical Engineering from National Tsing Hua University (NTHU), Hsinchu, Taiwan, in 2000. He was with the Department of Computer Science and Information Engineering, National Chung Cheng University, Taiwan, from 2000 to 2007. He is currently a Distinguished Professor with the Department of Electrical Engineering and the Institute of Communications Engineering, NTHU. He is also the Deputy Director of the NTHU AI Research Center. His research interests include image and video processing, computer vision, and video networking. Currently, he is Serving as an Associate Editor-in-Chief of \textsc{IEEE Transactions on Circuits and Systems for Video Technology}.
\end{IEEEbiography}





\end{document}